\documentclass[10pt,twocolumn,letterpaper]{article}

\usepackage{cvpr}              % To produce the CAMERA-READY version
\usepackage{cuted}
\usepackage{adjustbox}
\usepackage{bm}

\usepackage{multirow}
\usepackage{graphicx}
\usepackage{amsmath}
\usepackage{utfsym}
\usepackage{siunitx}    
\usepackage{amssymb} 
\usepackage{pifont}  
\usepackage{listings}
\usepackage{array} % 导言区引入

\usepackage{tabularx}
 \usepackage{booktabs}
\usepackage{tcolorbox}

\usepackage{threeparttable} % 用于带注释的表格
\usepackage{makecell}       % 用于单元格内换行

\usepackage[ruled,vlined,linesnumbered,noend]{algorithm2e}

\definecolor{cvprblue}{rgb}{0.21,0.49,0.74}
\usepackage[pagebackref,breaklinks,colorlinks,allcolors=cvprblue]{hyperref}

\def\paperID{} % *** Enter the Paper ID here
\def\confName{CVPR}
\def\confYear{2026}

\title{\textcolor{cvprblue}{Layer}-wise Instance \textcolor{cvprblue}{Bind}ing for Regional and Occlusion Control \\ in Text-to-Image Diffusion Transformers}

\author{Ruidong Chen$^{1}$~
        Yancheng Bai$^{2\ddag}$,~
        Xuanpu Zhang$^1$,~
        Jianhao Zeng$^1$,~
        Lanjun Wang$^{1}$,~ \\
        Dan Song$^{1}$,~
        Lei Sun$^2$,~
        Xiangxiang Chu$^2$,~
        Anan Liu$^{1\ast}$,~ \\
        $^1$Tianjin University,~
        $^2$Independent Researcher\\
}
\begin{document}

\twocolumn[{
\renewcommand\twocolumn[1][]{#1}
\maketitle
\begin{center}
    \vspace{-8mm}
    \captionsetup{type=figure}
    \includegraphics[width=0.95\textwidth]{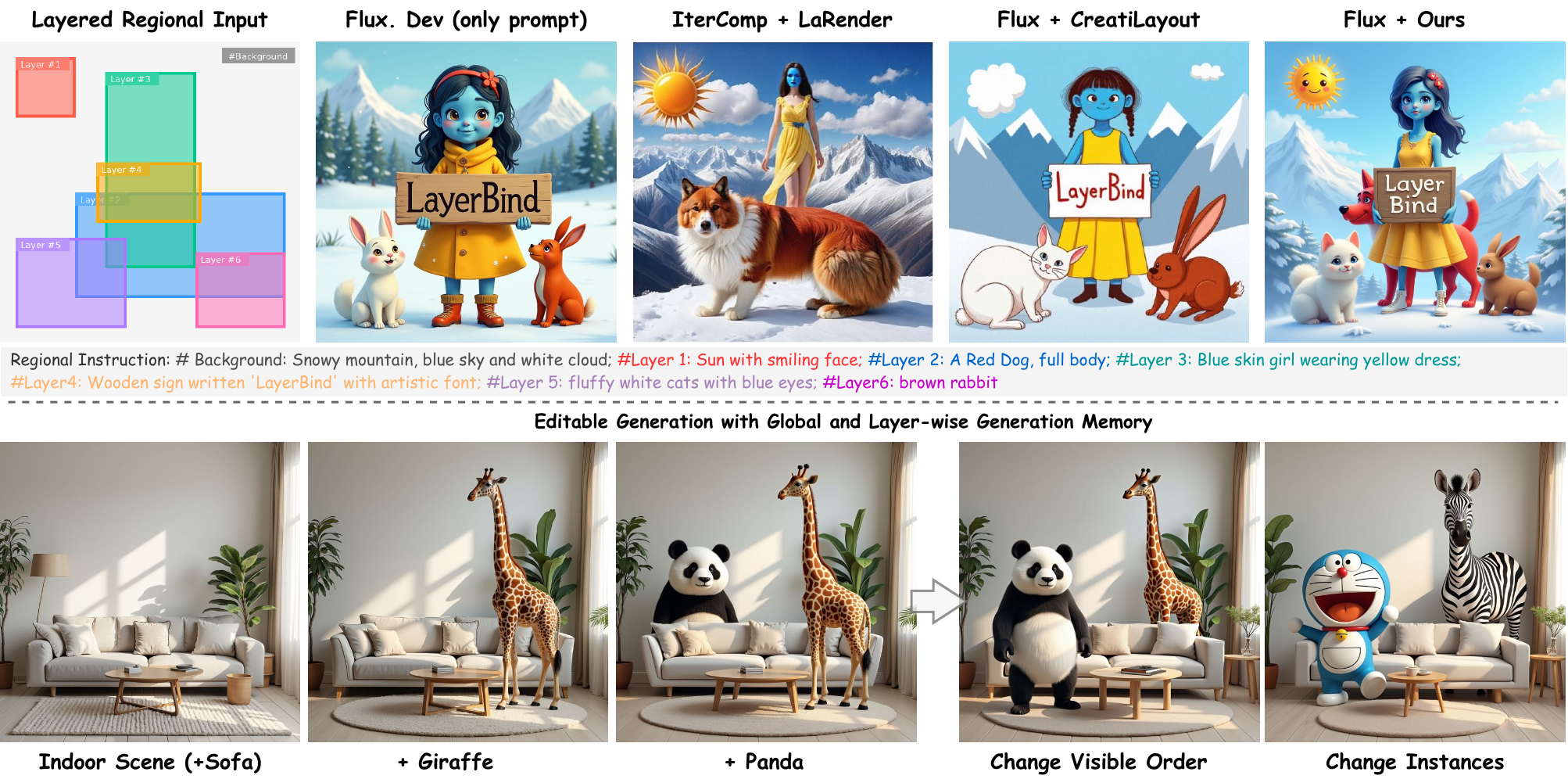}
    \vspace{-2mm}
    \captionof{figure}{
        We propose \textbf{LayerBind, a training-free strategy to empower text-to-image DiT models~\cite{bfl_flux1_dev_modelcard_2025,stability_sd3_5_large_modelcard_2025} with regional and occlusion controllability.} 
        % (\textbf{Top}) In contrast to prior training-free methods with weak layer control (LaRender~\cite{zhang2024itercomp,zhan2025larender}) and training-based methods that degrades image quality (CreatiLayout~\cite{bfl_flux1_dev_modelcard_2025,zhang2025creatilayout}), LayerBind produces high-quality images that respect the specified spatial layout and occlusion relations. 
        (\textbf{Top}) Compared to prior methods~\cite{zhan2025larender,zhang2025creatilayout}, LayerBind produces customized images that respect the specified spatial layout and occlusion relations without image quality degradation. 
        (\textbf{Bottom}) LayerBind is based on a context-sharing, region-branching strategy. This design inherently enables editable generation, allowing flexible modifications like changing per-region instances or visible orders.
    }
    \label{fig:fig1}
    % \vspace{-1mm}
\end{center}
}]

\renewcommand{\thefootnote}{}
% \footnotetext{
% % $^\dag$ Work done during the internship at AMAP, Alibaba Group.
% }
\footnotetext{
$^*$ Corresponding author; $^\ddag$ Project lead.
% : Lanjun Wang\{wang.lanjun\}@outlook.com,\\~An-An Liu\{anan0422\}@gmail.com
}

% \maketitle

% \setlength{\stripsep}{-21pt}

% \begin{strip}
%     \centering
%     % \vspace{-8mm}
%     \includegraphics[width=0.96\textwidth]{images/fig1.pdf}
    
%     \vspace{-2mm}
%     \captionof{figure}{
%         We propose \textbf{LayerBind, a training-free strategy to empower text-to-image DiT models~\cite{bfl_flux1_dev_modelcard_2025,stability_sd3_5_large_modelcard_2025} with regional and occlusion controllability.} 
%         % (\textbf{Top}) In contrast to prior training-free methods with weak layer control (LaRender~\cite{zhang2024itercomp,zhan2025larender}) and training-based methods that degrades image quality (CreatiLayout~\cite{bfl_flux1_dev_modelcard_2025,zhang2025creatilayout}), LayerBind produces high-quality images that respect the specified spatial layout and occlusion relations. 
%         (\textbf{Top}) Compared to prior methods~\cite{zhan2025larender,zhang2025creatilayout}, LayerBind produces customized images that respect the specified spatial layout and occlusion relations without image quality degradation. 
%         (\textbf{Bottom}) LayerBind is based on a context-sharing, region-branching strategy. This design inherently enables editable generation, allowing flexible modifications like changing per-region instances or visible orders.
%     }
%     \label{fig1}
%     % achieving high effectiveness in concept removal with minimal impact on generation quality.}
% \vspace{12mm}
% \end{strip}

\begin{abstract}
Region-instructed layout control in text-to-image generation is highly practical, yet existing methods suffer from limitations: (i) training-based approaches inherit data bias and often degrade image quality, and (ii) current techniques struggle with occlusion order, limiting real-world usability.
To address these issues, we propose \textbf{LayerBind}.
By modeling regional generation as distinct layers and binding them during the generation, our method enables precise regional and occlusion controllability.
Our motivation stems from the observation that spatial layout and occlusion are established at a very early denoising stage, suggesting that rearranging the early latent structure is sufficient to modify the final output.
Building on this, we structure the scheme into two phases: instance initialization and subsequent semantic nursing.
(1) First, leveraging the contextual sharing mechanism in multimodal joint attention, \textbf{Layer-wise Instance Initialization} creates per-instance branches that attend to their own regions while anchoring to the shared background. 
At a designated early step, these branches are fused according to the layer order to form a unified latent with a pre-established layout.
(2) Then, \textbf{Layer-wise Semantic Nursing} reinforces regional details and maintains the occlusion order via a layer-wise attention enhancement. 
Specifically, a sequential layered attention path operates alongside the standard global path, with updates composited under a layer-transparency scheduler.
LayerBind is training-free and plug-and-play, serving as a regional and occlusion controller across Diffusion Transformers. 
It also supports editable workflows, allowing for flexible modifications like changing instances or rearranging visible orders.
% Beyond generation, it natively supports editable workflows, allowing for flexible modifications like changing instances or rearranging visible orders.
% Both qualitative and quantitative 
Experimental results demonstrate LayerBind's effectiveness, highlighting its potential for creative applications.
Project page: \href{https://littlefatshiba.github.io/layerbind-page/}{https://littlefatshiba.github.io/layerbind-page}
\end{abstract}

\vspace{-5mm}
\section{Introduction}
\label{sec:intro}

Text-to-Image~(T2I) models have advanced rapidly, with Diffusion Transformers~(DiTs)~\cite{peebles2023scalable,sd3} emerging as the dominant architecture due to their strong scalability and high-fidelity generation quality.
To enhance controllability, region-instructed layout control~\cite{zhou2024migc,li2023gligen,zhang2025creatilayout,chen2025ragd,zheng2023layoutdiffusion} uses regional cues~(e.g., boxes or masks with associated instructions) to dictate instance placement and appearance.
% , has become a key technique.
This approach is valued for executing user-specified or LLM-parsed layout plans~\cite{chen2025ragd,yang2024mastering,feng2023layoutgpt} and has garnered widespread research attention.
However, most existing methods~\cite{li2023gligen,zheng2023layoutdiffusion,liang2025vodiff,zhou2024migc,phung2024grounded} are designed particularly for U-Net pipelines~\cite{sd14,sd3}, which are transferred poorly to DiTs, given their substantially different attention mechanisms, tokenization schemes, and model size. Consequently, research on DiT-native layout controllers remains limited.

Existing DiT-native layout controllers primarily follow two directions.
First, training-based methods, which fine-tune the DiT model~\cite{zhang2025creatilayout} or adopt layout adapters~\cite{hu2022lora,xiang2025instanceassemble,wu2025hybrid}. While these methods can achieve precise layout control, they introduce training data bias, causing image quality degradations (e.g. CreatiLayout~\cite{zhang2025creatilayout} in Fig.~\ref{fig:fig1}).
% noticeably degrade generation quality~(Fig.~1, top, Fig.~\ref{fig:compare1}).
%, and still struggle with complex spatial relationships 
Second, training-free methods, such as ~\cite{chen2025ragd,chen2024training,zhan2025larender}, conduct regional prompting to inject semantics into localized regions, often preserving the model's original generation quality. 
However, despite both having advantages, they fail to manage object occlusion, and often cause ``concept blending", where semantics from different regions erroneously fuse (Fig.~\ref{fig:compare1}). This highlights a critical and shared gap: \textbf{achieving robust control over both regional layout and occlusion with high-fidelity generation in DiTs} remains an unsolved problem.

% achieving precise and robust control over both regional layout and occlusion while maintaining high-fidelity generation remains an unsolved problem.

\begin{figure}[t]
    \centering
\includegraphics[width=0.9\linewidth]{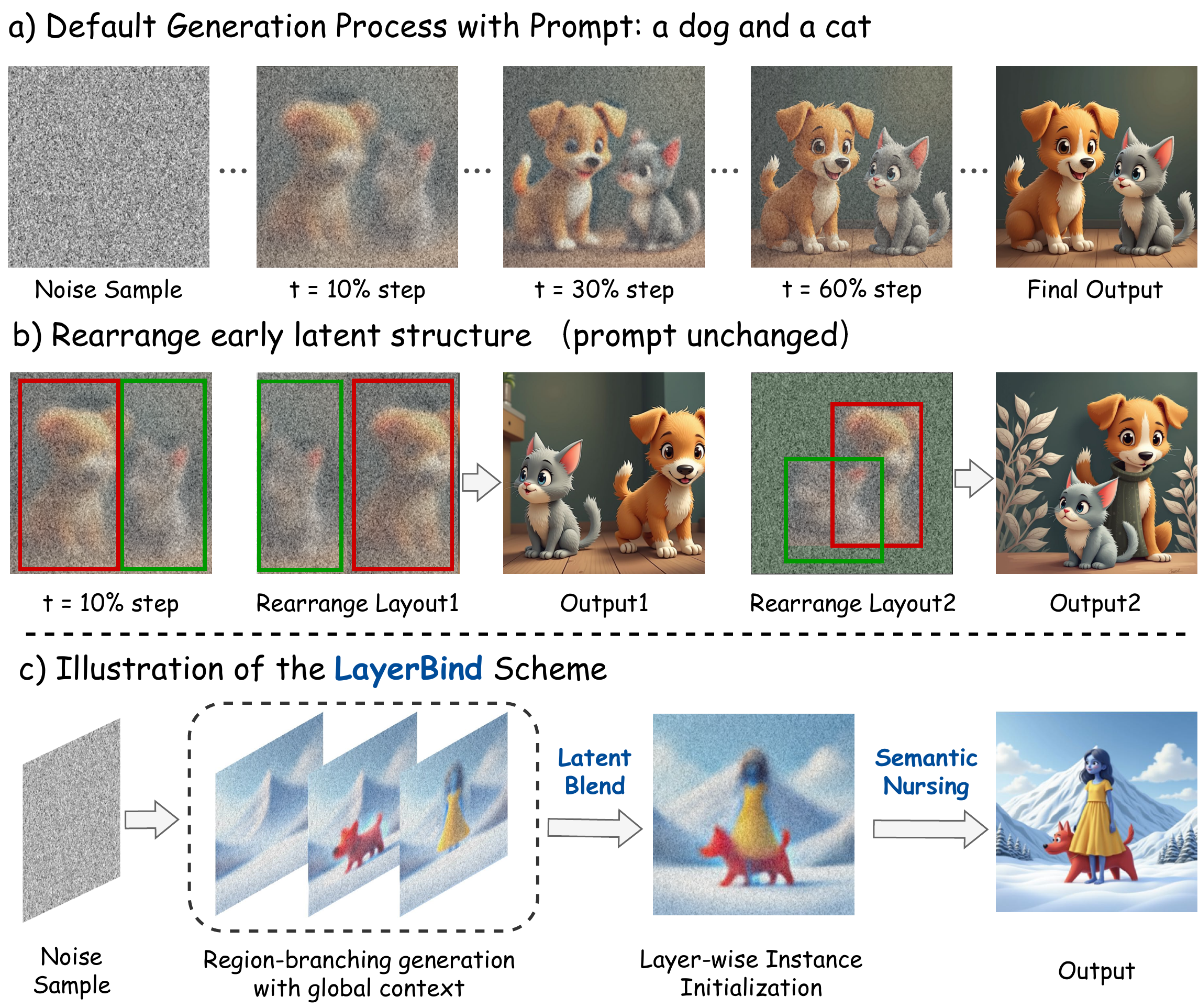}
    \caption{(a, b) Observation: simply rearranging the latent structure at an early step directly manipulates the final spatial layout and occlusion order. (c) Our LayerBind scheme: initializing the instance layout first, then conducting semantic nursing for instance detail while maintaining layout and occlusions.
    }
    \label{fig2}
    \vspace{-6mm}
\end{figure}

To address these limitations, we propose \textbf{LayerBind}, a training-free controller enabling precise regional and occlusion control for DiTs.
% Our approach is motivated by a key observation on denoising dynamics: the foundational layout and high-level structure are largely established at a very early denoising step~\cite{lipman2022flow,ssb} (Fig.~\ref{fig2}a).
% Building on this, we find that simply re-arranging the latent structure at this early stage directly manipulates the final spatial layout and occlusion ordering (Fig.~\ref{fig2}b).
Our approach is motivated by a key observation regarding the model's denoising dynamics: the foundational layout is rigidly established at a very early denoising step~\cite{ssb,lipman2022flow,mao2024lottery, kim2023leveraging} (Fig.~\ref{fig2}a), and through rearranging this early latent structure, we can directly modify the final layout and occlusion (Fig.~\ref{fig2}b).
This leads to our core motivation: \textbf{effective layout control should align with the model's intrinsic denoising dynamics}, rather than countering them at temporally misaligned stages. 
% LayerBind implements this principle by decoupling the task into two sequential phases (Fig.~\ref{fig2}c): 
% (1) an early-stage structural initialization to define layout and occlusion, and (2) a subsequent semantic nursing phase to refine instance details while maintaining layer integrity. 
% We implement this scheme via two core components: \textbf{Layer-wise Instance Initialization} and \textbf{Layer-wise Semantic Nursing}. 
LayerBind implements this principle by decoupling the region-instructed layout control into two sequential phases (Fig.~\ref{fig2}c):
(1) Layer-wise Instance Initialization, an early-stage process to initialize instances and define layout and occlusion, and (2) Layer-wise Semantic Nursing, a subsequent phase to refine instance details while maintaining occlusion integrity.

LayerBind implements this two-stage design as follows. 
(1) \textbf{Layer-wise Instance Initialization} first creates per-instance branch tokens from the initial latents, corresponding to each specified region. Leveraging the contextual sharing mechanism of multimodal joint-attention~(MM-Attention)~\cite{sd3}, these branches compute attention independently, allowing them to form distinct instances while adapting the shared background context. 
At a designated early step, these branches are fused into the global latent according to the desired layer order, either via direct latent merging or by using an optional foreground blend to composite the instances, thereby establishing the initial structured latent. 
% (a name inspired by Generative Semantic Nursing~\cite{chefer2023attend})
(2) \textbf{Layer-wise Semantic Nursing} then takes over after the merge. This stage performs semantic refinement via layer-wise local attention enhancements.
In each attention block, a standard global path runs, followed by a sequential layered path that processes each region and its instruction. 
A layer-transparency scheduler then manages these layered enhancements, ensuring regional semantics and layer relationships are progressively reinforced throughout the subsequent denoising process.

In summary, LayerBind is a training-free and plug-and-play controller that enables precise region and occlusion control for DiTs while preserving generation quality. 
Furthermore, its region-branching scheme inherently enables an editable generation.
% , as demonstrated in Fig. 1 (Bottom) and Fig.~\ref{fig:applications}. 
This design permits flexible modifications, such as changing per-region objects, altering occlusion order~(Fig.~\ref{fig:fig1}, Bottom), or even performing composited image editing with any image as background context~(Fig.~\ref{fig:applications}). 
Extensive quantitative and qualitative experiments validate LayerBind's state-of-the-art performance in complex layout and occlusion control.
\section{Related Work}
\label{sec:related}

% \noindent \textbf{Layout-to-Image Generation.} 
\subsection{Layout-to-Image Generation}
Layout-to-Image (L2I) generation~\cite{zheng2023layoutdiffusion, phung2024grounded,layoutguidance,liang2025vodiff,chen2024anyscene,dahary2024yourself,wang2024instancediffusion} aims to control the spatial layouts of synthesized images. 
A further extension, region-instructed L2I~\cite{feng2023layoutgpt,zhou2024migc,zhang2025creatilayout,xiang2025instanceassemble,li2023gligen,wang2024instancediffusion,jiang2025cmsl}, incorporates detailed regional information for finer-grained controllability. 
Early approaches diverged into three categories: (1) Training-based methods~\cite{zheng2023layoutdiffusion,zhou2024migc,li2023gligen} that finetune the model on layout inputs; (2) Latent optimization methods~\cite{liang2025vodiff,phung2024grounded,layoutguidance} that leverage spatial objectives to guide denoising; and (3) Seed-based methods~\cite{mao2024lottery, guo2024initno, lee2024groundit} that manipulate the initial noise for object placement.

% xie2023boxdiff

While these methods performed well on U-Net models (e.g., Stable Diffusion 1.5~\cite{sd14}), they struggle to adapt to DiT-based architectures, which feature larger parameter counts and fundamentally different attention mechanisms. Consequently, recent efforts have focused on solutions for these newer base models. 
Recent efforts include training-based adaptations for DiTs~\cite{zhang2025creatilayout,xiang2025instanceassemble,wu2025hybrid,lan2025flux,zeng2025eevee} and autoregressive models~\cite{he2025plangen,jin2025semantic}, as well as methods leveraging auxiliary modules like depth parsers~\cite{zhou2025dreamrenderer, zhou20253dis}. 
More prominently, training-free regional prompting methods~\cite{yang2024mastering,chen2025ragd,chen2024training} leverage the high-fidelity generation of pre-trained DiTs, often using LLMs as layout parsers, to achieve customized generation. 
However, as discussed in our introduction, these methods struggle with complex spatial relationships, particularly object occlusion. 
LayerBind builds upon this training-free paradigm, aiming to enhance layout precision and empower the model with robust occlusion control capabilities.

\subsection{Layer-wise Image Generation}
% \noindent \textbf{Image Generation with Layer Control.}
A distinct line of research, relevant to our ``layer-wise'' concept, employs explicit image layers to enhance generation. The most common approach~\cite{zhang2024transparent, huang2025dreamlayer, dalva2024layerfusion, tudosiu2024mulan} involves training models on RGBA decomposed object images, granting them the ability to generate foreground images with transparent backgrounds. Other works utilize pre-composed foregrounds as a prior to guide scene generation~\cite{khan2025composeanything, xu2025contextgen}, or use stored layer-wise memories to preserve unedited regions during multi-step image editing~\cite{kim2025improving}.
% , chen2024anyscene

A method particularly relevant to our task~(occlusion control) is LaRender~\cite{zhan2025larender}. Inspired by NeRF~\cite{mildenhall2021nerf} rendering, working on IterComp~\cite{zhang2024itercomp} and GLIGEN~\cite{li2023gligen} basemodel, LaRender replaces standard attention with a layer-ordered object rendering process, thereby explicitly modeling occlusion. However, this approach places stringent demands on its layer-wise prompts, often resulting in missing objects (Fig.~\ref{fig:fig1} and Fig.~\ref{fig:compare1}). 
In contrast, LayerBind's contextual sharing mechanism ensures each layer remains grounded in a shared background context, achieving more robust and accurate generations.
\section{Preliminaries}
\label{sec:pre}

\begin{figure*}[t]
    \centering
    \includegraphics[width=1\textwidth]{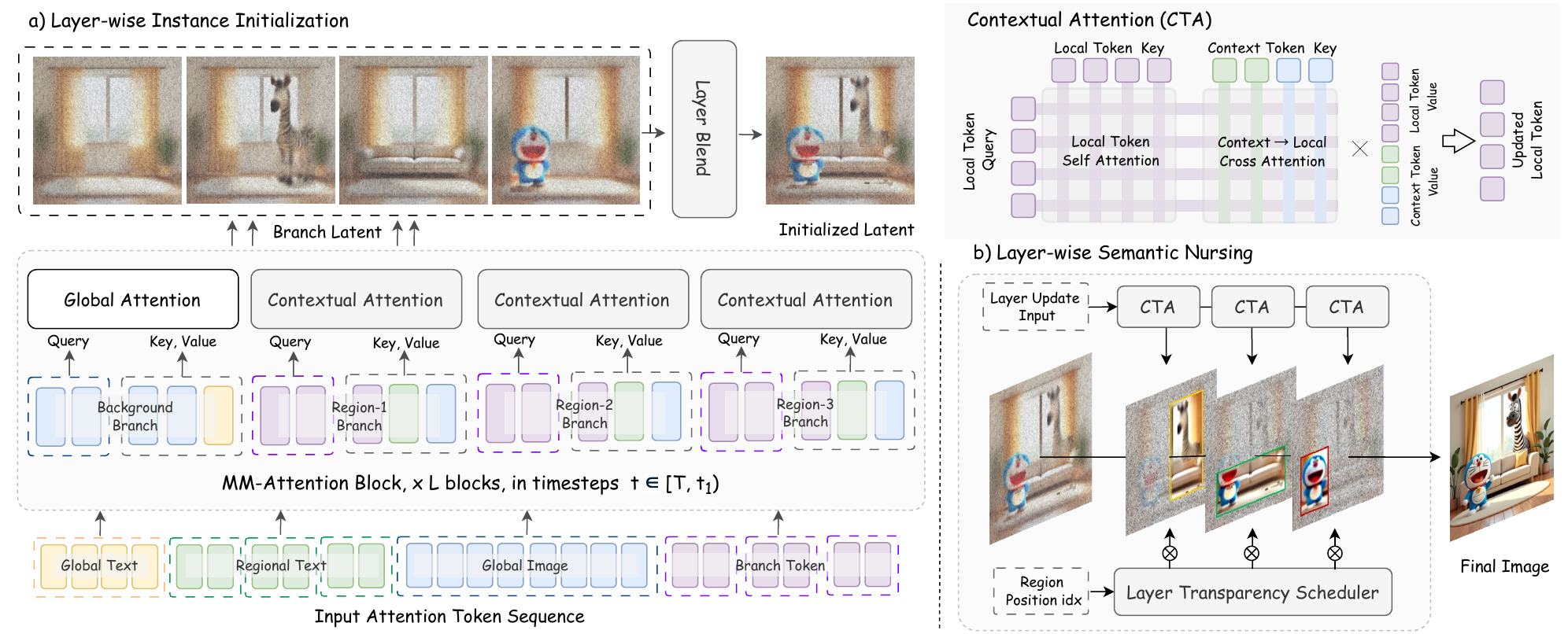}
    \captionof{figure}{Overview of the LayerBind pipeline. (a) \textbf{Layer-wise Instance Initialization} splits early denoising
    % ~(~$t\in[T,$))
    into background and instance branches.
    % (one background and three region branches in the figure)
    Each instance generates independently while sharing background context (via Contextual Attention, CTA, Eq.~\ref{eq:contextual_update}), then they are fused to form the initialized early latent. (b) \textbf{Layer-wise Semantic Nursing} reinforces following generation. It conduct layer-wise sequential CTA updates for each region, modulated by a Layer Transparency Scheduler, to refine instance details and maintain occlusions. \textit{Note: For simplicity, only image token updates are visualized; the detailed strategy will be described in the following sections.}
    }
    % t1到t2。 parallel， sequential
    \label{fig:pipeline}
    \vspace{-5mm}
    % \vspace{10mm}
\end{figure*}

To introduce LayerBind, this section reviews the relevant preliminaries.

\noindent \textbf{Rectified Flow Model}. Mainstream DiTs~\cite{sd3,bfl_flux1_dev_modelcard_2025} are based on rectified-flow models~\cite{sd3}. 
During denoising inference, at time \(t\), denote the data sample as $\boldsymbol{x}$ and condition as $\boldsymbol{y}$, a network predicts a velocity field \(v_{\theta}(\boldsymbol{x}_t, t \mid \boldsymbol{y})\) that transports samples along a linear noise-to-data path.
Sampling integrates the ODE using an explicit Euler solver on a linearly spaced timestep grid, yielding the denoising trajectory:
\begin{equation}
% \small
\boldsymbol{x}_{k-1}=\boldsymbol{x}_k \;+\; (t_{k-1}-t_k)\, v_{\theta}(\boldsymbol{x}_k, t_k \mid \boldsymbol{y}).
\
\end{equation}
This trajectory makes each state the initial condition for all subsequent updates, so simple rearrangements applied early deterministically propagate through the entire trajectory, directly supporting our early-binding motivation for layout and occlusion control.

\noindent\textbf{Multimodal Diffusion Transformers~(MM-DiT)}.
Modern DiTs~\cite{sd3} unify textual and visual tokens via a single joint attention operator, computing self-attention over a unified sequence to enable bidirectional context sharing.
Let text and image tokens be \(T\in\mathbb{R}^{N_T\times d_T}\) and \(I\in\mathbb{R}^{N_I\times d_I}\).
They are mapped to a shared \(d\)-dimensional space to produce queries ($Q_\mathrm{T}, Q_\mathrm{I}$), keys ($K_\mathrm{T}, K_\mathrm{I}$), and values ($V_\mathrm{T}, V_\mathrm{I}$).
Joint attention \(A_{\text{joint}}\) is then calculated by concatenating tokens from both modalities:
\begin{equation}
\label{eq:joint-attention}
\small
A_{\text{joint}}(Q,K,V) = \text{Softmax} \left( \frac{[Q_\mathrm{T} \oplus Q_\mathrm{I}] [K_\mathrm{T} \oplus K_\mathrm{I}]^\top}{\sqrt{d}} \right)[V_\mathrm{T} \oplus V_\mathrm{I}],
\end{equation}
where \(\oplus\) denotes concatenation along the sequence dimension.
This formulation allows all tokens to freely attend to each other, enabling powerful cross-modal reasoning.

\noindent\textbf{Localized Attention with Contextual Update}.
The flexibility of joint attention (Eq.~\ref{eq:joint-attention}) allows for updating local tokens by selectively constructing $Q$, $K$, and $V$, enabling fine-grained information flow~\cite{helbling2025conceptattention,chen2025normalized,kimseg4diff,zhang2025group}.
Specifically, given local tokens ($T/I_{\text{local}}$) and context tokens ($T/I_{\text{context}}$), we can update the former using them as queries ($Q_{\text{local}}$), while concatenating both to form keys and values ($K=[K_{\text{local}} \oplus K_{\text{context}}]$, $V=[V_{\text{local}} \oplus V_{\text{context}}]$).
With Eq.~\ref{eq:joint-attention}, we define this operation, which we term \textbf{``Contextual Attention''} (Fig.~\ref{fig:pipeline}), as:
\begin{equation}
\label{eq:contextual_update}
\small
\mathcal{A}_{\text{update}}(Q_{\text{local}}, [K_{\text{local}} \oplus K_{\text{context}}], [V_{\text{local}} \oplus V_{\text{context}}]).
\end{equation}
To simplify notation and the explanation of proposed LayerBind, we abbreviate this operation as $\hat{e}_{\text{out}} \leftarrow \mathcal{A}_{\text{update}}(e_{\text{query}}, 
e_{\text{context}})$, where $e_{\text{context}}$ may be a concatenation of multiple contexts, such that $ e_{\text{context}}=[e_{\text{ctx\_1}},e_{\text{ctx\_2}}, \dots]$.
This functional representation maintains consistency with joint attention and is mathematically equivalent to attention masking~\cite{zhou2025dreamrenderer,chen2025ragd, chen2024training, wei2025freeflux}, yet improves efficiency and clarifies the information flow for local token updates.

\section{Method}
\label{sec:Method}

\subsection{Problem Definition}
\label{sec:problem_definition}

This study focuses on the task of \textit{``region-instructed layout and occlusion control in DiT models"}.
Following prior work on region-instructed layout control~\cite{chen2025ragd,yang2024mastering, chen2024training, zhang2025creatilayout,feng2023layoutgpt, khan2025composeanything}, the inputs are precisely user-defined or LLM-parsed structured inputs, including: a background prompt ($T_{\text{bg}}$) for the initialization stage, a full scene prompt ($T_{\text{scene}}$) for the subsequent nursing stage, and a set of $N$ layered regional inputs.
Each layer $i$ includes a regional prompt ($T_{\text{reg}}^{(i)}$) and a corresponding spatial cue ($C^{(i)}$) (e.g., box or mask).
The layer index $i$ explicitly defines the occlusion order, from farthest ($i=1$) to nearest ($i=N$).
% camera
This cue $C^{(i)}$ corresponds to a set of token indices $idx^{(i)}$ in the DiT token sequence.

Given these inputs, the goal is to generate an image that satisfies three key requirements:
\begin{itemize}
    \item \textbf{Layout \& Occlusion:} Strict adherence to the spatial cues $C^{(i)}$ and the \textbf{pre-defined} $i$-th layer occlusion order.
    \item \textbf{Regional Fidelity:} Faithful semantic control for each instance $T_{\text{reg}}^{(i)}$ without concept blending.
    \item \textbf{Global Harmony:} High-fidelity quality and coherent composition, preserving the base model's capabilities.
\end{itemize}
To meet these requirements, we introduce LayerBind (Fig.~\ref{fig:pipeline}).
Our approach decouples this task into two sequential stages: (1) \textbf{Layer-wise Instance Initialization} to first establish the layout and (2) \textbf{Layer-wise Semantic Nursing} to subsequently refine details and maintain integrity. The following sections will detail each component.

\subsection{Layer-wise Instance Initialization}
\label{sec:initialization}

This stage operates during the initial phase of denoising. Let $T$ be the maximum diffusion timestep (e.g., $T=1000$) and $S$ be the total number of discrete inference steps.
This initialization stage is active for the first $\eta_1$ ratio of inference steps.
This defines a timestep threshold $t_1$, such that this phase runs during the interval $t \in [T, t_1)$.
At step $t_1$, the branches are fused to form the initialized latent.

\noindent \textbf{Branch Construction}.
At the initial denoising step $t=T$, we construct each branch $B^{(i)}$ by directly copying from the global latent $I$ at the specified indices:
% \vspace{-5pt}
\begin{equation}
\label{eq:branch_construct}
B^{(i)}(t=T) \leftarrow I(t=T)[idx^{(i)}].
\end{equation}
Inside each DiT block, these latents ($I$ and $B^{(i)}$) are mapped to their corresponding \textit{embeddings}. As illustrated in Fig.~\ref{fig:pipeline}(a), the attention blocks operate on four main types: the global image embedding ($e_I$), the instance branch ($e_B^{(i)}$), the global text ($e_{T_{\text{bg}}}$), and the regional text($e_{T_{\text{reg}}}^{(i)}$).
Crucially, $e_B^{(i)}$ also inherits the RoPE position embeddings~\cite{su2024roformer} from $e_I[idx^{(i)}]$.
Based on the ODE sampling properties (Sec.~\ref{sec:pre}), this shared starting point ensures $I$ and all $B^{(i)}$ share the same underlying noise structure, naturally promoting global consistency even as their semantic paths diverge.

\noindent \textbf{Branch Updates with Contextual Attention}.
The core of the branch update policy is to create a bidirectional binding between each instance branch $e_B^{(i)}$ and its corresponding regional text $e_{T_{\text{reg}}}^{(i)}$, while grounding both in the shared visual background context.
We define this background context $e_{I_{\text{bg}}}^{(i)}$ as the set of global image tokens excluding the $i$-th region (i.e., $e_{I_{\text{bg}}}^{(i)} = e_I[\sim idx^{(i)}]$).
In parallel, the main embedding $e_I$ and $e_{T_{bg}}$ are updated via standard joint attention.
Using the abbreviated Eq.~\ref{eq:contextual_update}, the instance branch update is:
% notation of 
\vspace{-3pt}
\begin{equation}
\small
\label{eq:branch_update}
\hat{e}_B^{(i)} \leftarrow \mathcal{A}_{\text{update}}(
    e_B^{(i)},~ [e_{I_{\text{bg}}}^{(i)}, e_{T_{\text{reg}}}^{(i)}]
).
\end{equation}
This allows the branch to adapt to the background content (via $e_{I_{\text{bg}}}^{(i)}$), while simultaneously ingesting its semantic guidance (from $e_{T_{\text{reg}}}^{(i)}$).
Symmetrically, the regional text is also updated to reflect the emerging visual features:
\begin{equation}
\small
\label{eq:text_update}
\hat{e}_{T_{\text{reg}}}^{(i)} \leftarrow \mathcal{A}_{\text{update}}(
    e_{T_{\text{reg}}}^{(i)},~ [e_B^{(i)}, e_{I_{\text{bg}}}^{(i)}]
).
\end{equation}
Together, Eq.~\ref{eq:branch_update} and \ref{eq:text_update} create a localized feedback loop, refining both the instance's semantics and its textual guidance simultaneously.

\begin{figure}[t]
    \centering
\includegraphics[width=0.9\linewidth]{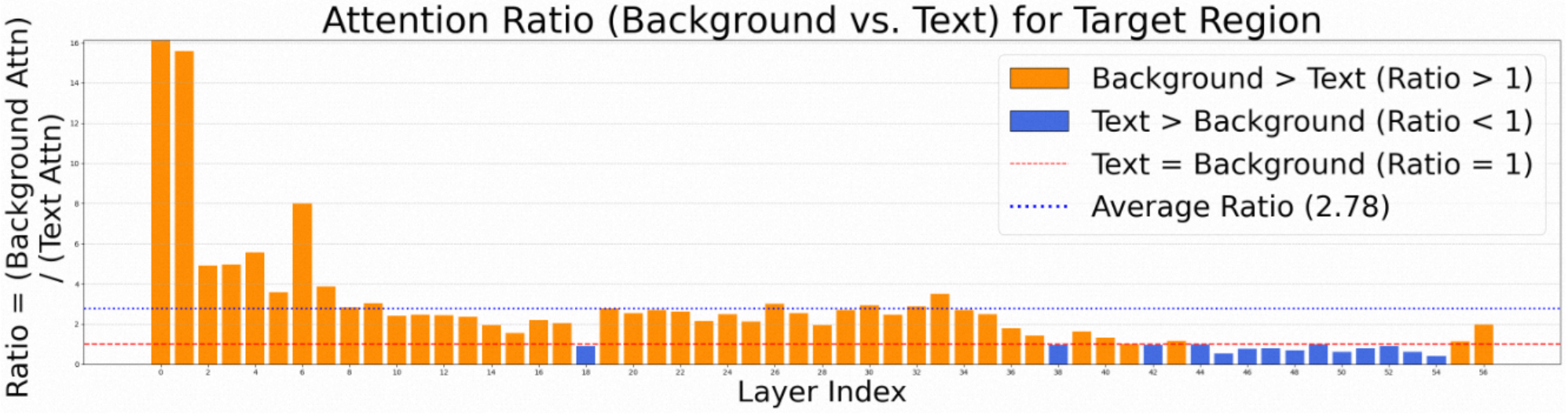}
    \caption{
    Attention response weights of foreground to background and text across different FLUX~\cite{bfl_flux1_dev_modelcard_2025} layers. We select layer 0~\cite{wei2025freeflux,avrahami2025stable} and layers with strong text response for hard instance binding. 
    % This enhances semantic injection while avoiding excessive disruption to background adaptation. 
    More analysis is presented in the Appendix~\ref{sup:method}.
    }
    \label{fig:ratio}
    \vspace{-5mm}
\end{figure}

\noindent \textbf{Hard Binding and Reverse Adaptation}.
A known failure mode during instance initialization is ``modality competition''~\cite{lv2025rethinking}, where strong background semantics can overwhelm the weaker regional text signal $e_{T_{\text{reg}}}^{(i)}$, causing small objects to be ignored.
To mitigate this, we leverage the observation that certain DiT blocks exhibit significantly stronger text responses~\cite{wei2025freeflux,avrahami2025stable} (Fig.~\ref{fig:ratio}).
In these ``text-dominant'' blocks, we employ a hard binding and reverse adaptation policy: First, a hard binding forces the instance branch $e_B^{(i)}$ to update exclusively from itself and its guiding text, severing the link to the background:
\vspace{-5pt}
\begin{equation}
\small
\label{eq:branch_update_hard}
\hat{e}_B^{(i)} \leftarrow \mathcal{A}_{\text{update}}(
    e_B^{(i)},~ [e_{T_{\text{reg}}}^{(i)}]
).
\end{equation}
Second, a reverse adaptation ensures boundary harmony. We force the background regions $e_{I_{\text{bg}}}^{(i)}$ to adapt to the branch region $e_B^{(i)}$, and to ``emptying out'' space for it:
\begin{equation}
\small
\label{eq:bg_update}
\hat{e}_{I_{\text{bg}}}^{(i)} \leftarrow \mathcal{A}_{\text{update}}(
    e_{I_{\text{bg}}}^{(i)},~ [e_{T_{\text{bg}}}, e_B^{(i)}]
).
\end{equation}
In practice, this asymmetric update~(Eq.~\ref{eq:bg_update}) is implemented with a structured attention mask.
This strategy ensures small instances receive sufficient textual guidance, while the reverse adaptation maintains a seamless blend between the instance and the scene.

% alpha i
\noindent \textbf{Layer-wise Branch Blending}.
\label{sec:layer_blend}
At the designated blend step $t_1$, the $N$ instance branches $B^{(i)}$ are sequentially fused into the global latent $I$ according to the occlusion order.
We employ a conditional strategy: unoccupied (bottom) layers are directly merged ($I[idx^{(i)}] \leftarrow B^{(i)}$), while for occluding (top) layers, an optional foreground alpha mask $\alpha_f^{(i)}$ is estimated~(details in Appendix~\ref{sup:method}) to prevent background interference and improve edge quality.
These occluding layers are then composited via layer order:
\vspace{-5pt}
\begin{equation}
\label{eq:layer_blend}
I[idx^{(i)}] \leftarrow \alpha_f^{(i)} \cdot B^{(i)} + (1 - \alpha_f^{(i)}) \cdot I[idx^{(i)}].
\end{equation}
This yields the final initialized latent with an explicit layout structure.

\subsection{Layer-wise Semantic Nursing}
\label{sec:nursing}

Following initialization, this stage maintains the established layout and reinforces regional details during $t \in (t_1, t_2]$, where $t_2$ is determined by a ratio $\eta_2$ of the total inference steps $S$. This stage utilizes the full scene prompt $T_{\text{scene}}$ as the global text condition.
In each attention block, a standard global attention $\hat{e}_I^{\text{global}}$ is computed (using $e_I$ and $e_{T_{\text{scene}}}$).
In parallel, for each layer $i$, we compute a local attention enhancement $\hat{e}_{\text{local}}^{(i)}$ for its image region $e_{I_{reg}}^{(i)} = e_I[idx(i)]$, and concurrently update its regional text $e_{T_{\text{reg}}}^{(i)}$:
\vspace{-5pt}
\begin{equation}
\small
\label{eq:nursing_update}
\hat{e}_{\text{local}}^{(i)} \leftarrow \mathcal{A}_{\text{update}}(
    e_{I_{reg}}^{(i)},~ [e_{T_{\text{reg}}}^{(i)},e_I]
),
\end{equation}
\begin{equation}
\small
\label{eq:nursing_update_text}
\hat{e}_{T_{\text{reg}}}^{(i)} \leftarrow \mathcal{A}_{\text{update}}(
    e_{T_{\text{reg}}}^{(i)},~ [e_{I_{reg}}^{(i)}, e_{T_{\text{scene}}}]
).
\end{equation}
To compose the final embedding $\hat{e}_I^{\text{out}}$, these local enhancements $\hat{e}_{\text{local}}^{(i)}$ are sequentially blended onto the global result $\hat{e}_I^{\text{global}}$ via a transparency scheduler.
Following the occlusion order (bottom $i=1$ to top $N$), we define the base layer as $\hat{e}_{\text{comp}}^{(0)} = \hat{e}_I^{\text{global}}$ and compute the final output via an iterative update:
\vspace{-10pt}
\begin{equation}
\small
\label{eq:transparency_scheduler}
\begin{split}
    \hat{e}_{\text{comp}}^{(i)} &= (1 - \alpha_o^{(i)}) \cdot \hat{e}_{\text{comp}}^{(i-1)} + \alpha_o^{(i)} \cdot \hat{e}_{\text{local}}^{(i)},
\end{split}
\end{equation}
where $\alpha_o^{(i)} = \beta \cdot M^{(i)}$, with $\beta$ is the opacity factor and $M^{(i)}$ is the binary mask for region $i$.
% and $\odot$ denotes token-wise multiplication.
This iterative compositing ensures the semantics of top layers robustly overwrite bottom layers in overlapping regions.

\begin{figure*}[t]
    \centering
    \includegraphics[width=1\textwidth]{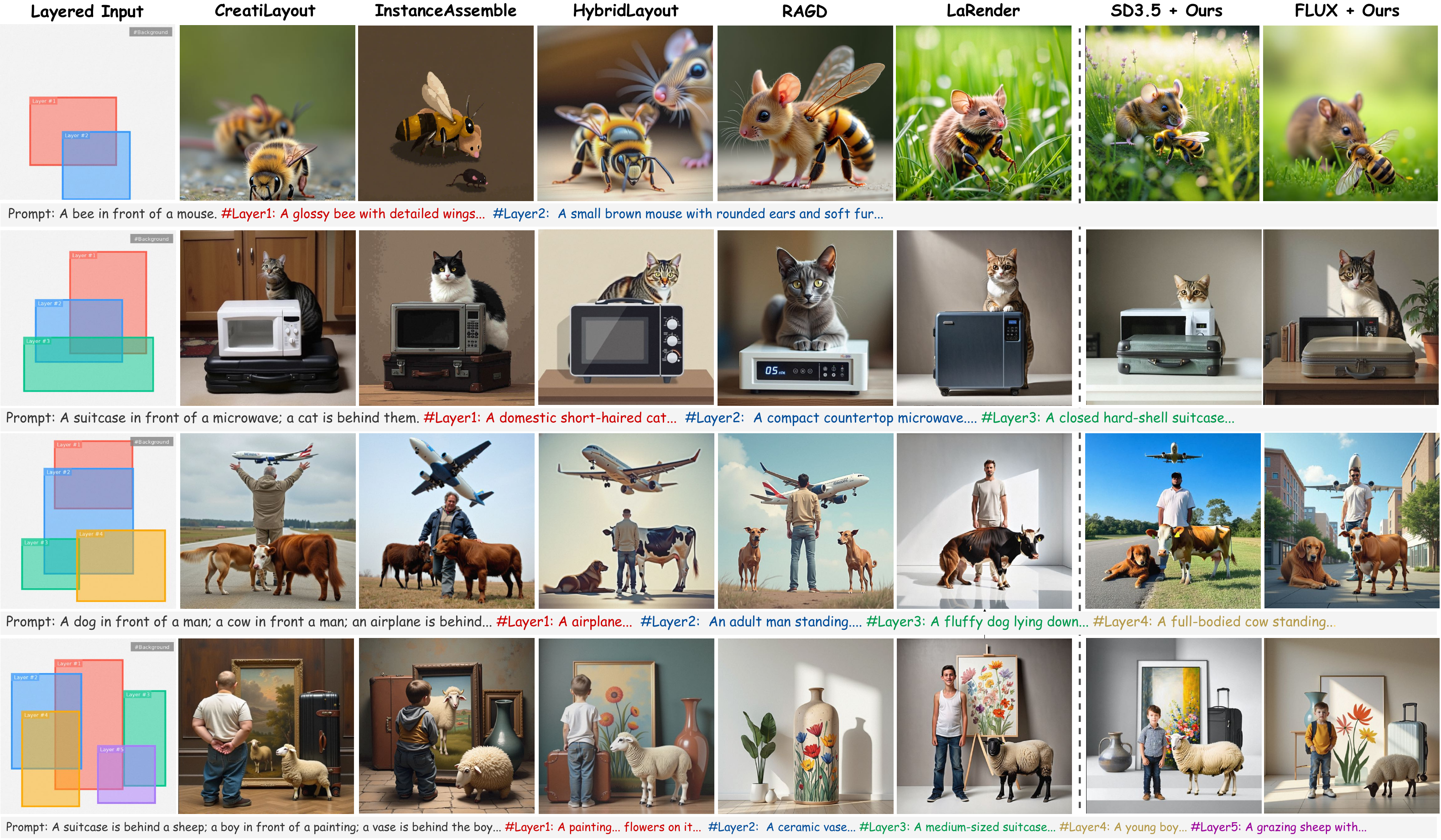}
    \captionof{figure}{Visualization of occlusion control abilities. Compared to the previous methods, LayerBind achieves more precise layer-wise control, avoiding errors such as instance neglect and concept blending. More visualizations are available in the Appendix~\ref{sup:vis}} 
    \label{fig:compare1}
    \vspace{-5mm}
    % \vspace{10mm}
\end{figure*}

\section{Experiments}

% \subsection{Implementation Details}
% \label{sec:implementation}

% We implement our method primarily on the FLUX.1-dev~\cite{bfl_flux1_dev_modelcard_2025} model. 
% We adopt its default generation settings: $S=28$ total inference steps and a guidance scale of 3.5.
% Correspondingly, $\eta_1$ for Phase 1 (Sec.~\ref{sec:initialization}) is set to 0.2, and $\eta_2$ for Phase 2 (Sec.~\ref{sec:nursing}) is set to 0.7 by default. The opacity factor $\beta$ (Eq.~\ref{eq:transparency_scheduler}) is set to 0.8.
% Besides Flux, we also validate the adaptability of LayerBind with two other mainstream DiTs: SD3.5~\cite{stability_sd3_5_large_modelcard_2025} and Qwen-Image.
% Partial results are presented in this section, with full implementation details deferred to the Appendix.

\subsection{Evaluation Settings}

\noindent \textbf{Implementation Details}. 
We implement LayerBind on two mainstream DiT models: FLUX.1-dev~\cite{bfl_flux1_dev_modelcard_2025} and SD3.5 Large~\cite{stability_sd3_5_large_modelcard_2025}.
For evaluation, we adopt the default generation settings for both models (e.g., inference steps and guidance scale) and apply LayerBind's two-stage strategy.
Specifically, $\eta_1$ for Phase 1 (Sec.~\ref{sec:initialization}) is set to 0.2 for FLUX and 0.25 for SD3.5, while $\eta_2$ for Phase 2 (Sec.~\ref{sec:nursing}) is set to 0.7 by default.
The opacity factor $\beta$ (Eq.~\ref{eq:transparency_scheduler}) is set to 0.7.

\noindent \textbf{Baselines}. We compare LayerBind against SOTA methods that accept layouts with regional instructions as input.
These include training-based approaches~\cite{li2023gligen,zhang2025creatilayout,wu2025hybrid,xiang2025instanceassemble} and training-free approaches~\cite{chen2024training,chen2025ragd,zhan2025larender}.
Among these, LaRender is the only method focused on occlusion control; we compare against both its GLIGEN~\cite{li2023gligen} and IterComp~\cite{zhang2024itercomp} implementations.
For fair comparison, we use their official implementations with identical seeds and inputs for all generations and evaluations.

\noindent \textbf{Evaluation Benchmarks}.
To validate the practical application of LayerBind for high-quality customized image generation, we evaluate both occlusion control and T2I alignment tasks.
For occlusion control, we use the 3D-spatial subset of T2I-CompBench~\cite{huang2023t2i}.
To address its limitation to simple two-object relations, we additionally construct BindBench for complex occlusions among 3-5 objects~(Appendix~\ref {sup:dataset}).
% After layout annotation and filtering, we obtain 800 and 200 samples for these two benchmarks, respectively.
For the general T2I task, we use the attribute bindings, spatial, numeracy, and complex subsets of T2I-CompBench.
Lacking publicly released layout annotations, we follow~\cite{zhang2025creatilayout,khan2025composeanything,chen2025ragd,chen2024training} and employ an LLM (e.g., GPT-5-mini~\cite{gpt5}) for layout parsing~(Appendix~\ref {sup:layout_parser}).

% with Chain-of-Thought prompting for layout parsing.

\noindent \textbf{Metrics}.
We employ a multi-perspective evaluation for occlusion control: 1) UniDet-Depth~\cite{huang2023t2i} measures relative object depth between two objects; 2) CLIP Score (global/local) assesses text-image consistency at both scene and instance levels; 3) O$_{VQA}$ and L$_{Acc/VQA}$, build upon VQAScore~\cite{lin2024evaluating}, quantifies the perceptual score of occlusion relations and layout faithfulness; 4) HPS is reported to assess generation quality~\cite{wu2023human}. For general T2I alignment, we use the official T2I-CompBench metrics. Please refer to the Appendix~\ref {sup: metrics} for full details of our implementation and evaluation protocols.

\begin{table*}[t]
\centering
\renewcommand{\arraystretch}{0.98}
\resizebox{1\textwidth}{!}{%
\begin{tabular}{lccc ccccc cccccc c} 
\toprule % 替换 \hline
\multicolumn{1}{c}{\multirow{2}{*}{Method}} & \multicolumn{1}{c}{\multirow{2}{*}{Model}} & \multicolumn{1}{c}{\multirow{2}{*}{\makecell{Training-\\free}}} & \multicolumn{5}{c}{T2ICompBench-3D~\cite{huang2023t2i}} & \multicolumn{6}{c}{BindBench (Ours)} & \multicolumn{1}{c}{\multirow{2}{*}{\makecell{Inference \\ Speed~(s)}}} \\ 
\cmidrule(lr){4-8} \cmidrule(lr){9-14}
\multicolumn{1}{c}{} & \multicolumn{1}{c}{} & \multicolumn{1}{c}{} & UniDet$\uparrow$ & CLIP$_G\uparrow$ & CLIP$_L\uparrow$ & O$_{VQA}$ $\uparrow$ & HPS $\uparrow$ & CLIP$_G\uparrow$ & CLIP$_L\uparrow$ & L$_{Acc}\uparrow$ & L$_{VQA}\uparrow$ & O$_{VQA}$ $\uparrow$ & HPS $\uparrow$ & \multicolumn{1}{c}{} \\ \midrule % 替换 \hline
 &SD3.5  & - & 40.06 & 34.00 & - & 40.54 & 28.37 & 36.33 & - & - & - & 23.06 & 30.93 & 20.96 \\
& FLUX & - & 37.97 & 33.17 & - & 37.56 & 29.66 & 36.08 & - & - & - & 18.86 & 30.98 & 23.12 \\ \midrule
InstanceDiffusion~\cite{wang2024instancediffusion} & SD-2.1 & \usym{2717} & \underline{41.53} & 33.02 & 28.37 & 44.73 & 26.42 & 35.55 & 27.71 & 76.46 & 40.63 & 24.09 & 26.20 & 16.71~(+308\%) \\
GLIGEN-XL~\cite{li2023gligen} & SD-XL & \usym{2717} & 35.13 & 33.13 & 26.97 & 41.22 & 26.00 & 35.03 & 25.85 & 74.36 & 25.32 & 24.43 & 26.29 & \textbf{10.52}~(+4\%) \\
CreatiLayout*~\cite{zhang2025creatilayout} & FLUX & \usym{2717} & 39.37 & \underline{33.67} & 27.79 & 57.03 & 27.38 & \underline{36.28} & 26.60 & 85.18 & 40.99 & 43.62 & 28.73 & 16.75~(+148\%) \\
HybridLayout~\cite{wu2025hybrid} & FLUX & \usym{2717} & 41.33 & 32.85 & 26.97 & 47.55 & 26.43 & 35.45 & 26.14 & 78.30 & 43.45 & 34.10 & \underline{29.20} & 78.75~(+240\%) \\
InsAssem~\cite{xiang2025instanceassemble}& FLUX & \usym{2717} & 31.45 & 32.29 & 25.97 & 40.66 & 25.37 & 34.94 & 25.08 & 67.24 & 27.43 & 30.21 & 26.42 & \underline{24.86}~(+7\%) \\ 
CreatiDesign~\cite{zhang2026creatidesign}& FLUX & \usym{2717} & 41.00 & \textbf{33.80} & 24.58 & 38.60 & \textbf{28.56} & \textbf{36.34} & 23.01 & 61.12 & 25.59 & 26.26 & \textbf{29.79} & 75.05~(+224\%) \\ \midrule %
RAGD~\cite{chen2025ragd} & FLUX & \usym{2713} & 30.13 & 32.21 & 27.61 & 31.22 & 26.64 & 30.43 & 26.97 & 36.61 & 20.81 & 1.82 & 22.80 & 62.14~(+168\%) \\
LaRender~\cite{zhan2025larender} & GLIGEN & \usym{2713} & 39.41 & 32.40 & 27.52 & 41.95 & 25.73 & 34.30 & 27.18 & 60.52 & 38.53 & 27.24 & 25.83 & 12.49~(+23\%) \\
LaRender~\cite{zhan2025larender} & IterComp & \usym{2713} & 37.52 & 33.14 & 27.85 & 35.96 & 27.37 & 32.47 & 25.77 & 60.52 & 42.62 & 17.72 & 26.27 & 13.10~(+29\%) \\ \midrule %
LayerBind~(Ours) & SD3.5 & \usym{2713} & 41.37 & 33.02 & \underline{28.49} & \textbf{65.78} & \underline{28.36} & 35.01 & \underline{27.25}  & \underline{87.57} & \underline{59.73} & \underline{48.03} & 29.03 & 29.39~(+40\%) \\ 
LayerBind~(Ours) & FLUX & \usym{2713} & \textbf{44.97} & 33.12 & \textbf{28.54} & \underline{59.49} & 28.25 & 35.72 & \textbf{27.86} & \textbf{92.18} & \textbf{64.81} & \textbf{52.55} & \underline{29.66} & 30.11~(+30\%) \\ \bottomrule
\end{tabular}
} % \resizebox 的结束括号
\caption{Quantitative comparison for occlusion control, measuring: depth relationship (UniDet), T2I alignment (CLIP-G/L), Layout alignment (L$_{Acc/VQA}$), occlusion perception score (O$_{VQA}$), and image quality (HPS). Inference speed includes the percentage of overhead introduced by the controller relative to the base model. * evaluated at 512x512 resolution; all other methods are at 1024x1024.}
\label{tab:compare1}
\vspace{-10pt}
\end{table*}

\subsection{Main Results}

\begin{figure}[t]
    \centering
\includegraphics[width=1\linewidth]{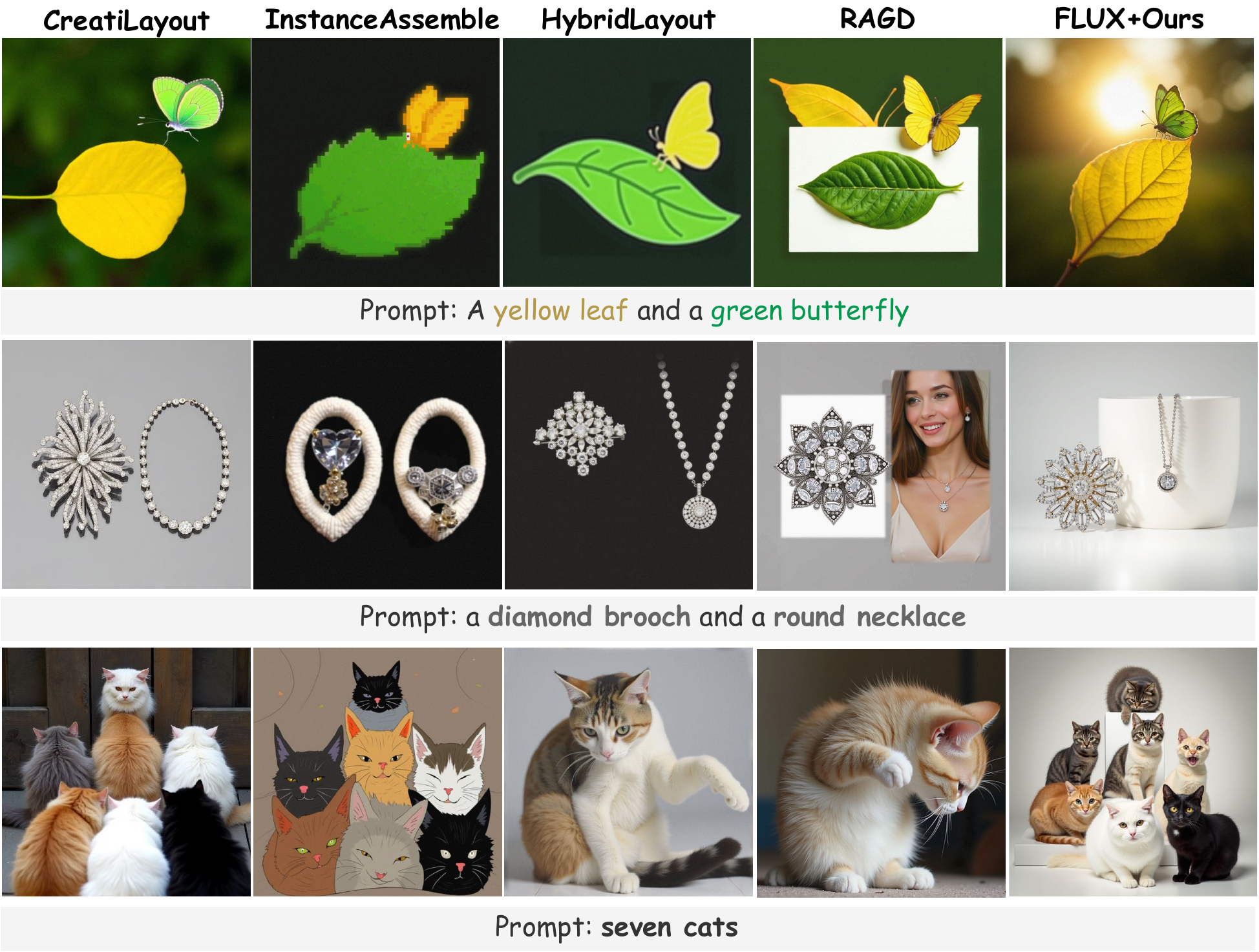}
    \caption{Visualization results on T2I alignment evaluations. LayerBind can serve as a plug-and-play layout controller for improving T2I alignment ability without quality degradation.
    }
    \label{fig:compare2}
    \vspace{-3mm}
\end{figure}

\noindent \textbf{Qualitative Results}
Figs.~\ref{fig:compare1} and ~\ref{fig:compare2} show qualitative comparisons of LayerBind on occlusion control and T2I alignment tasks, respectively.
LayerBind achieves superior alignment with user inputs across both tasks, realizing precise layout and occlusion control while minimizing impact on generation quality.
In contrast, while HybridLayout~\cite{wu2025hybrid} and LaRender~\cite{zhan2025larender} similarly adopt a divide-and-conquer strategy for regions, they struggle during region fusion, often leading to concept blending and missing instances. RAGD~\cite{chen2025ragd} maintains good image quality, but has difficulty handling complex overlapping layouts.
Among training-based methods, CreatiLayout~\cite{zhang2025creatilayout} demonstrates the most stable spatial layout capability (likely because it is fully fine-tuned rather than LoRA-based), yet it still fails to handle complex occlusion scenarios.
These results underscore the potential of LayerBind as a practical, training-free DiT layout controller for real-world creative applications.

\begin{table}
\centering
\resizebox{1\columnwidth}{!}{%
\begin{tabular}{lcccccc}
\toprule
\multirow{2}{*}{} & \multicolumn{3}{c}{Attribute Binding} & \multirow{2}{*}{Spatial} & \multirow{2}{*}{Numeracy} & \multirow{2}{*}{Complex} \\
\cmidrule(lr){2-4} %
& Color & Shape & Texture & & & \\
\midrule
FLUX~\cite{bfl_flux1_dev_modelcard_2025} & 77.53 & 60.16 & 69.64 & 39.09 & 59.81 & 37.01 \\
CreatiLayout~\cite{zhang2025creatilayout} & 76.94 & 59.92 & 73.45 & 60.33 & \textbf{71.51} & 37.45 \\
InstanceAssemble~\cite{xiang2025instanceassemble} & 71.64 & 55.13 & 62.15 & \underline{64.38} & 56.90 & 38.33 \\
HybridLayout~\cite{wu2025hybrid} & \underline{84.15} & \textbf{68.82} & \textbf{77.31} & 63.39 & 64.57 & 40.15 \\
RAGD~\cite{chen2025ragd} & 80.39 & 60.16 & 70.85 & 51.93 & 53.76 & \textbf{43.77} \\
LayerBind+FLUX (Ours) & \textbf{84.80} & \underline{66.48} & \underline{75.69} & \textbf{70.63} & \underline{70.93} & \underline{41.43} \\
\bottomrule
\end{tabular}
}
\caption{The quantitative evaluation results of T2I alignment tasks.}
\label{tab:compare2}
\vspace{-5mm}
\end{table}

\noindent \textbf{Quantitative Results}.
Table~\ref{tab:compare1} presents the quantitative results for occlusion control, where LayerBind (on both FLUX and SD3.5) achieves state-of-the-art performance.
On T2ICompBench-3D, LayerBind surpasses all competitors in the depth-occlusion metric (UniDet), demonstrating its ability to generate more natural scene depth.
LayerBind's VQA Score is also notably high across both benchmarks. This advantage is most pronounced on our challenging BindBench, where the performance of most methods degrades sharply, while LayerBind remains robust, proving its reliability in handling complex occlusions.
Furthermore, LayerBind attains the highest HPS score, confirming it best preserves image quality.
Regarding inference speed, LayerBind's efficient local attention mechanism is significantly faster than other region-partitioned generation methods (e.g., RAGD~\cite{chen2025ragd}, HybridLayout~\cite{wu2025hybrid}).

Additionally, Table~\ref{tab:compare2} shows the T2I alignment results.
Beyond basic attribute binding tasks, LayerBind's mechanism enables superior performance on difficult Numeracy and Complex tasks, significantly outperforming all existing methods.
This indicates that LayerBind is practical not only in occlusion control but also in general T2I generation.

\begin{table}[t]
\centering
\renewcommand{\arraystretch}{0.9}
\begin{adjustbox}{width=1\columnwidth}
\begin{tabular}{
>{\centering\arraybackslash}m{1.1cm}
>{\centering\arraybackslash}m{1.1cm}
|cccc}
\toprule
$HB$ & $LSN$ & CLIP$_G\uparrow$ & CLIP$_L\uparrow$ & VQAScore $\uparrow$ & HPS $\uparrow$ \\
\midrule
\usym{2717} & \usym{2717} & 34.95 & 26.82 & 38.36 & 28.27 \\
\usym{2717} & \usym{2713} & 34.73 & 26.90 & 43.65 & 28.64 \\
\usym{2713} & \usym{2717} & \textbf{35.78} & 27.80 & 50.98 & 29.64 \\
\usym{2713} & \usym{2713} & 35.72 & \textbf{27.86} & \textbf{52.55} & \textbf{29.66} \\
\bottomrule
\end{tabular}
\end{adjustbox}
\caption{The quantitative ablation results of applying Hard Binding~(Sec.~\ref{sec:initialization}, HB) and Layer-wise Semantic Nursing~(Sec.~\ref{sec:nursing}, LSN) on BindBench dataset.}
\label{tab:ablation}
\vspace{-2mm}
\end{table}

\begin{figure}[t]
    \centering
\includegraphics[width=1\linewidth]{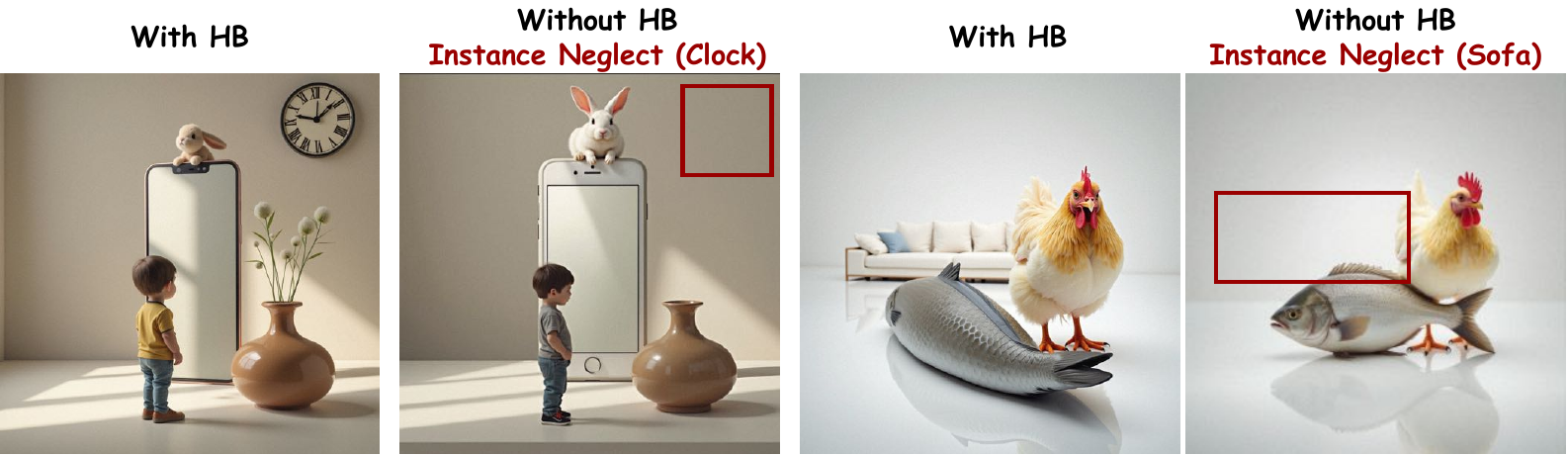}
    \caption{Visualization of effect of Hard Binding. It prevents instances from being ignored due to modality competition~\cite{lv2025rethinking}.
    }
    \label{fig:ablation1}
    \vspace{-2mm}
\end{figure}

\begin{figure*}[t]
    \centering
    \includegraphics[width=1\textwidth]{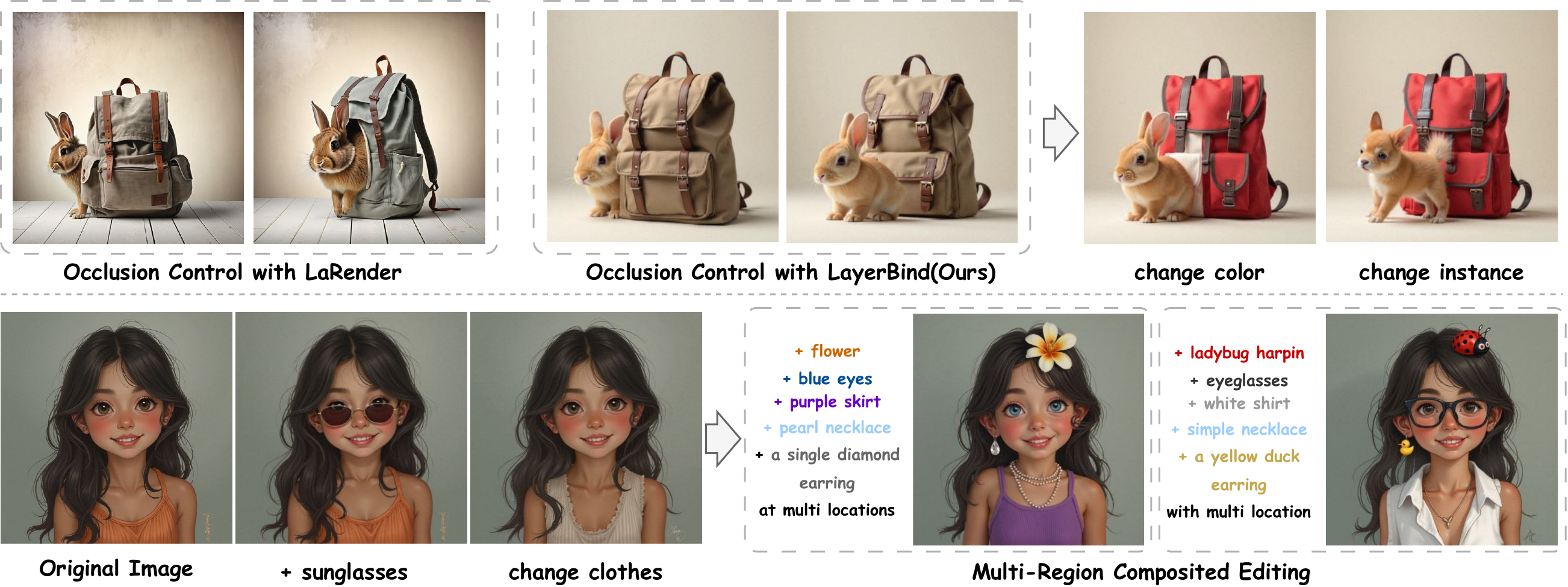}
    \captionof{figure}{Applications. Top) As also shown in Fig.1, LayerBind supports flexible occlusion control and instance modifications. Bottom) Treat an original generation as background context and branching edit instructions. LayerBind also achieves composited image edits.} 
    \label{fig:applications}
    \vspace{-4mm}
    % \vspace{10mm}
\end{figure*}

\begin{figure}[t]
    \centering
    \includegraphics[width=0.9\linewidth]{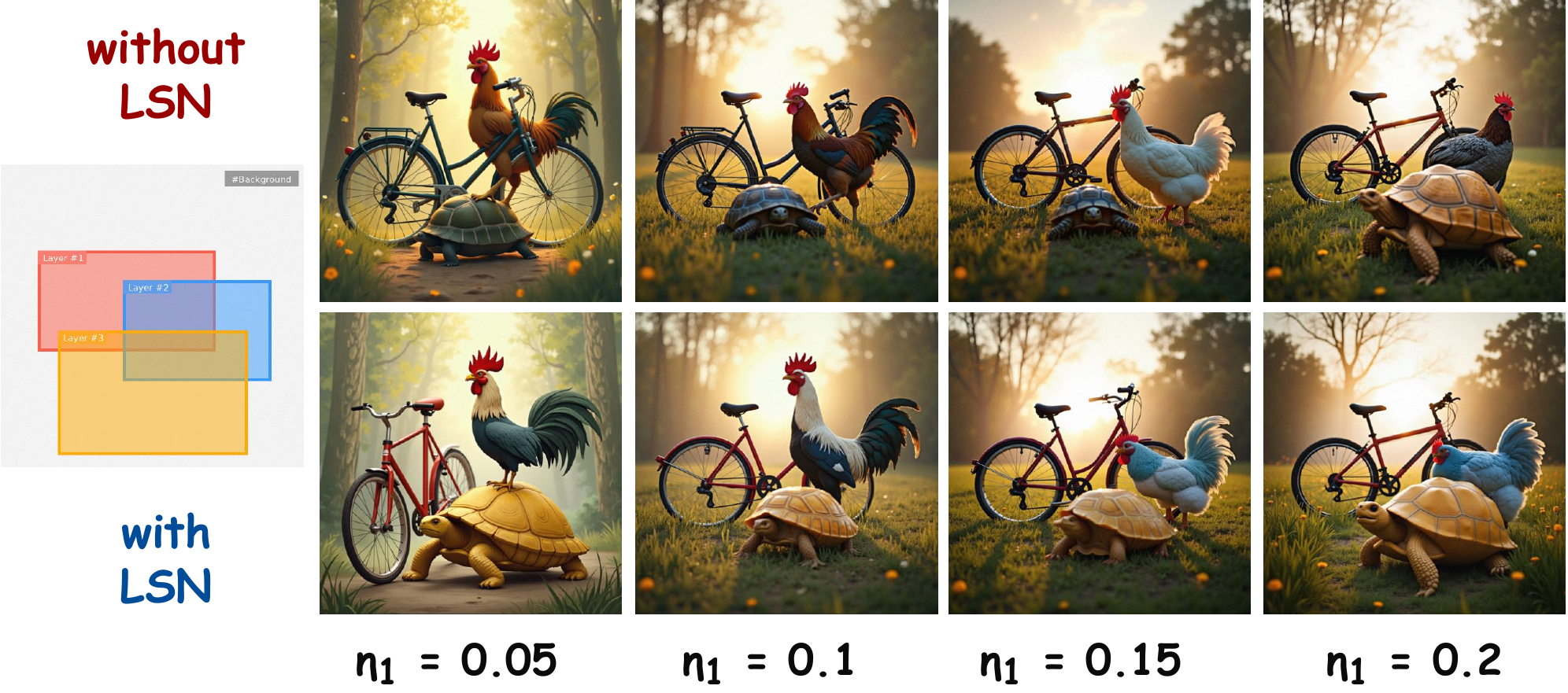}
    \caption{Visualization of effects of different $\eta_1$ with LSN strategy. To illustrate the details refinement of LSN, we add color attributes to each region~(e.g., golden turtle, blue chicken, red bicycle).
    }
    \label{fig:ablation2}
    \vspace{-6mm}
\end{figure}

\subsection{Ablation Analysis}

\noindent \textbf{Module Effectiveness}. 
Table~\ref{tab:ablation} presents the ablation results for our core components on BindBench (using $\eta_1=0.2$).
The results show that Hard Binding (HB) plays a decisive role in the overall occlusion success rate (VQAScore). Layer-wise Semantic Nursing (LSN) primarily refines regional details while also improving image quality.
Fig.~\ref{fig:ablation1} illustrates the qualitative impact of HB, which is critical in two scenarios: (1) small objects that are otherwise ignored by global attention, and (2) objects visually similar to the background that fail semantic initialization.
Overall, HB strengthens local semantic injection, thereby improving both the instance initialization and the final result.

\noindent \textbf{Effects of different $\eta_1$ with LSN}.
As shown in Fig.~\ref{fig:ablation2}, the two phases play complementary roles.
First, $\eta_1$ controls the structural initialization: the top row (without LSN) demonstrates that a higher $\eta_1$ is required to establish the correct structure.
The effectiveness of LSN is twofold: it helps maintain the layout and occlusions, and strengthens regional details.
The bottom row demonstrates that LSN successfully injects the correct color attributes even when the initial structure is poor.
Therefore, LSN serves as a complementary bridge. We can use a moderate $\eta_1$ to establish structure, avoiding the instance-background over-decoupling caused by an overly high $\eta_1$ (see Appendix~\ref{sup:ablation}), while LSN ensures the refinement of semantic details, guaranteeing overall harmony.

% For more ablation analysis, please refer to Appendix~\ref {sup:ablation}

% \noindent \textbf{Efficiency}

\subsection{Applications}

As illustrated in Fig.~\ref{fig:fig1} and Fig.~\ref{fig:applications}, LayerBind's region-branching mechanism supports flexible applications.
First, for layout and occlusion-control tasks, LayerBind serves as a \textbf{flexible layout controller} for precise regional-aware generation~(Fig.~\ref{fig:compare1},~\ref{fig:compare2}).
Furthermore, LayerBind's decoupling of the background and instance generation processes allows the initialization phase to act as a ``shared memory'' for the generation, enabling flexible operations such as \textbf{instance modifications}~(Fig.~\ref{fig:fig1}) and \textbf{occlusion order modifications}~(Fig.~\ref{fig:applications}), all while maintaining consistency in unaffected regions.
This inherent editability also extends to \textbf{composited image editing}~(Fig.~\ref{fig:applications}).
For example, a standard generation trajectory (initialized from the same noise or obtained via inversion~\cite{wang2024taming}) can serve as the background context.
LayerBind can then process regional edit instructions in separate branches and merge them back into the original trajectory.
This enables a powerful, multi-instruction editing capability that preserves irrelevant content, a promising application for interactive content creation.
In summary, LayerBind is highly extensible, and we discuss its more applications in Appendix~\ref{sup:app}.

\section{Conclusion and Limitations}

We propose \textbf{LayerBind}, a training-free regional and occlusion controller for text-to-image DiTs. LayerBind decouples the task into two stages: Layer-wise Instance Initialization to establish an early latent with pre-defined layout and occlusion, and Layer-wise Semantic Nursing to further refine details and maintain layouts.
Extensive experiments validate LayerBind's superior performance in complex occlusion control and general T2I alignment tasks, along with application analysis, showing its potential for interactive image customization and image editing applications.

\noindent \textbf{Limitations}. Despite its superior performance, LayerBind has certain limitations.
For example, some cases exhibit incomplete object generation, instance-background over-decoupling, or poor adherence to unreasonable layouts.
We showcase these failure cases in Appendix~\ref{sup:limitation} and discuss potential solutions for repair, in order to provide a more comprehensive reference for future work.

{
    \small
    \bibliographystyle{ieeenat_fullname}
    \bibliography{main}
}
\appendix
\clearpage
\setcounter{page}{1}
\maketitlesupplementary

\noindent This supplementary material is organized as follows:
\begin{itemize}
    \item Sec.~\ref{sup:method} details optional modules of LayerBind, including the vital block selection for the Hard-Binding and the implementation details of the Layer Blending mechanism.
    
    \item Sec.~\ref{sup:implement} provides detailed introductions of the experimental setup, including the strategy for layout parsing, the dataset construction, and the evaluation metrics.
    
    \item Sec.~\ref{sup:ablation} provides supplementary experiment analyses, including efficiency and further module ablations.
    
    \item Sec.~\ref{sup:app} further discusses applications of LayerBind, including the implementation details for composited image editing and its compatibility with external visual adapters.
    
    \item Sec.~\ref{sup:limitation} address the limitations of LayerBind. We showcase representative failure cases and discuss potential solutions for repair. 
    
    \item Sec.~\ref{sup:vis} provides more visualization of generation results.
\end{itemize}

\section{Optional Modules of LayerBind}
\label{sup:method}

\subsection{Vital Block Selection for Hard-Binding}

As discussed in prior work~\cite{lv2025rethinking}, "modality competition" can cause instances to be suppressed by strong background contexts.
To address this, we leverage the observation that different DiT blocks exhibit varying sensitivities to text~\cite{wei2025freeflux,avrahami2025stable}.
We suggest that forcing ``text-dominant" blocks to focus exclusively on their primary strength (semantic injection) offers a natural, minimally disruptive intervention for instance initialization.

\begin{figure}[t]
    \centering
\includegraphics[width=0.9\linewidth]{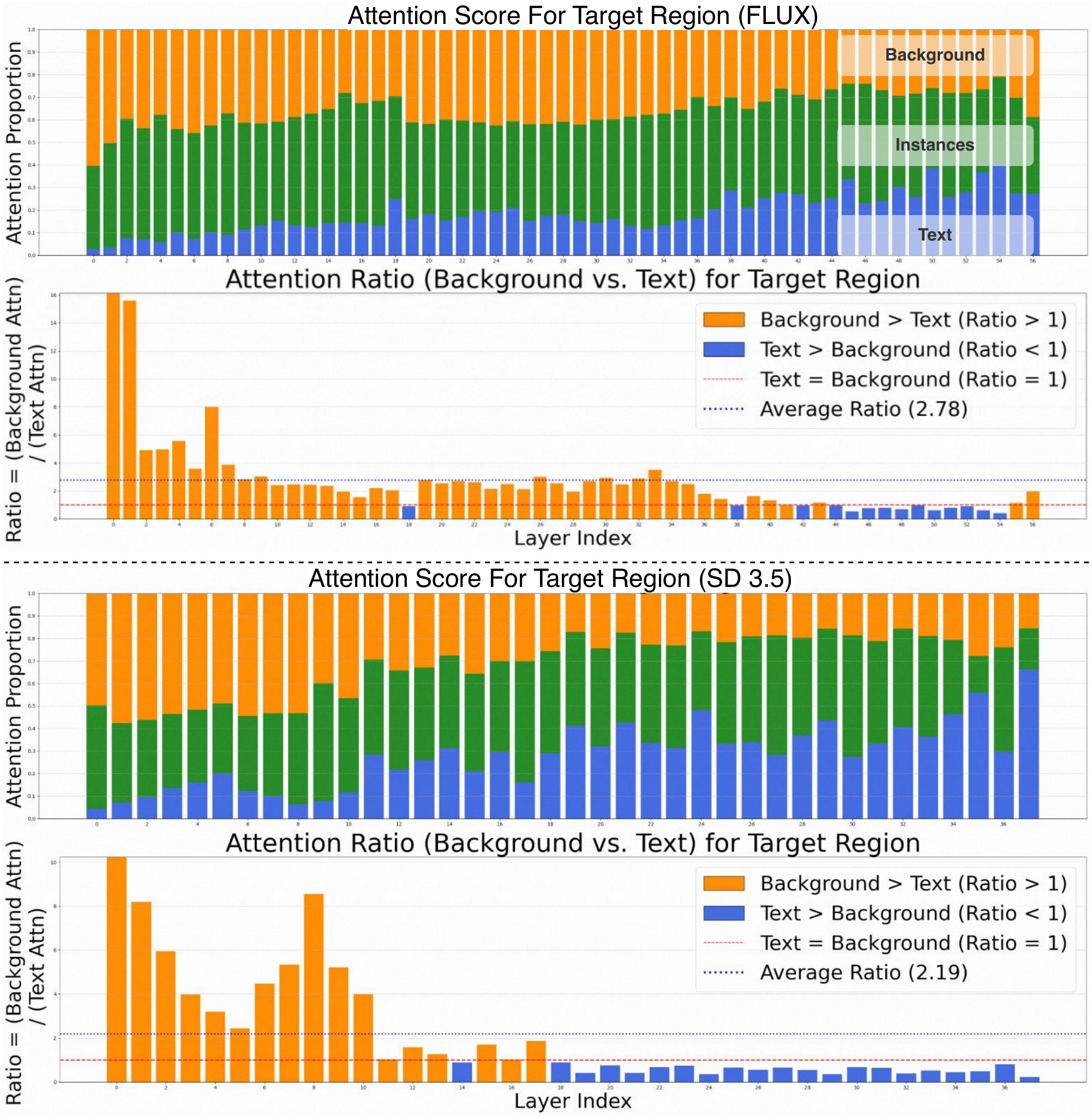}
    \caption{
    Attention response weights of foreground to background, self, and text tokens across different attention blocks~(in FLUX and SD3.5). }
    \label{fig:ratio_all}
    \vspace{-5mm}
\end{figure}

To identify these vital blocks, we empirically analyze attention maps recorded during generation.
Using segmentation tools on multiple runs (20 prompts, 5 seeds), we extract foreground masks and calculate the response intensity of foreground tokens (as queries) to different token types, averaged over the first 20\% of steps.
The results (Fig.~\ref{fig:ratio_all}) reveal a consistent trend in both FLUX and SD3.5: while foreground self-attention remains high across all layers, responses to background versus text tokens diverge significantly.
Early blocks are visually dominant, whereas mid-to-late blocks become progressively more text-responsive. 
Based on these findings, the selection is twofold:
\begin{enumerate}
    \item We always select Layer 0, uniquely critical for establishing initial semantic binding~\cite{wei2025freeflux,avrahami2025stable}.
    \item We select the 2 most text-responsive blocks from early-mid stages and the 6 most text-responsive blocks from late stages.
\end{enumerate}
This empirical selection balances semantic injection strength with maintaining image generation quality.
Specifically, we select blocks $[0, 15, 18, 42, 45, 48, 50, 53, 54]$ for FLUX (partially aligning with~\cite{wei2025freeflux,avrahami2025stable}) and $[0, 11, 14, 19, 21, 24, 29, 32, 34]$ for SD3.5.
Restricting Hard Binding to these vital blocks maximizes semantic injection while minimizing disruption to overall quality, as applying it to excessive blocks degrades performance.

% \subsection{Layer Blending Module}

\subsection{Layer Blending Module}

As described in Sec.~\ref{sec:layer_blend}, we composite occluding layers using a foreground-aware alpha mask $\alpha^{(i)}$. Since the branch $B^{(i)}$ and global latent $I$ share the same background context, we can estimate this mask directly from their difference. 
First, we compute a robust saliency map $Z$ by normalizing the difference using the local background variance ($\sigma_{\text{bg}}$), estimated from the Median Absolute Deviation (MAD) of a surrounding background area with the estimated region:
\begin{equation}
\small
    Z = \text{Smooth}\left( \left\| \frac{B^{(i)} - I[idx^{(i)}]}{\sigma_{\text{bg}} + \epsilon} \right\|_2^\gamma \right)
\end{equation}
where $I[idx^{(i)}]$ is the global latent in the instance region and $\gamma$ is a correction factor, default with 0.9.
This coarse saliency map $Z$ is then refined into a spatially smooth alpha mask by solving a Screened Poisson equation via an iterative solver:
\begin{equation}
\small
    \alpha_{k+1} = \frac{1}{4+\lambda} \left( \sum_{p \in \mathcal{N}(\alpha_k)} \alpha_p + \lambda Z \right)
\end{equation}
where $\lambda$ balances the data term $Z$ against the smoothness term.
After convergence, we apply an optional Otsu's thresholding~(a classic, unsupervised algorithm for automatic foreground thresholding) to explicitly separate the foreground region.
Finally, we perform morphological reconstruction to fill any internal holes, ensuring the instance appears contiguous and enhancing the blend's realism.

\section{Extended Evaluation Details}
\label{sup:implement}

\subsection{LLM-based Layout Parser}
\label{sup:layout_parser}

As outlined in Sec.~\ref{sec:problem_definition}, consistent with prior works~\cite{yang2024mastering,chen2025ragd,zhang2025creatilayout,khan2025composeanything}, we employ an LLM to generate layout plans when precise user-specified layouts are unavailable.
However, LayerBind has specific input requirements that generic parsers neglect: a decoupled background prompt and an explicit occlusion order.
To address this, we design a specialized prompt engineering strategy using a state-of-the-art LLM (e.g., GPT-5-mini~\cite{gpt5}), as visualized in Fig.~\ref{fig:llmprompt}.
The parsing process leverages Chain-of-Thought (CoT) prompting and is divided into two distinct stages:

\begin{enumerate}
    \item \textbf{Spatial reasoning:} The LLM first engages in a planning phase to analyze the input caption. It deduces the overall scene structure, ensuring logical spatial relationships and reasonable object placement before committing to coordinates.
    \item \textbf{Structured output:} We constrain the LLM to output a JSON. Crucially, this schema requires the model to explicitly generate a \texttt{"background\_prompt"} for context isolation and assign an \texttt{"order"} index to each instance based on its layer index, thereby rigorously defining the occlusion hierarchy.
\end{enumerate}

To ensure stability and adherence to this schema, we utilize in-context learning by providing few-shot examples (as shown at the bottom of Fig.~\ref{fig:llmprompt}), enabling the model to learn the correct patterns for layout generation.

Furthermore, this parser serves as the foundation for our benchmark construction.
Although prior works~\cite{yang2024mastering, zhang2025creatilayout, chen2025ragd, khan2025composeanything} have utilized layout generation for evaluating T2I-CompBench, none have publicly released their ground-truth layout annotations.
This absence prevents a fair, unified comparison across different methods.
Therefore, we use this standardized parser to re-annotate the dataset, ensuring a consistent evaluation ground for all baselines.

\tcbset{
  colback=gray!5,
  colframe=gray!40,
  boxrule=0.3pt,
  arc=2pt,
  outer arc=2pt,
  left=4pt,right=4pt,top=4pt,bottom=4pt
}

\begin{figure}[t]
\centering
\begin{tcolorbox}[title={Layer Parser Prompt (Simplified)}]
\small
% \begin{verbatim}
You are a layout annotator for text-to-image generation.

Goal: From an input text prompt, produce ONLY a JSON object with:

- planning, rewritten prompt,

- background prompt,

- instances list: instances with isolated captions, 1024×1024 bounding boxes, and depth order.

Rules (Detailed rules are attached after each item):

1. Plan the layout: ...

2. Rewrite the caption: ...

3. Instance extraction: ...

4. Generate detailed caption for each instance: ...

5. Relative depth annotation: ...

6. Bounding box layout annotation: ...

7. Generate background prompt for overall scene: ...

Input:
$<$INPUT\_PROMPT$>$

OUTPUT JSON schema:

\{

  \ \ ``planning": ``string",
  
  \ \ ``rewritten\_prompt": ``string",
  
  \ \ ``background\_prompt": ``string",
  
  \ \ ``elements": [
  
\ \  \{
    
\ \ \ \ ``region\_prompt": ``string",
      
\ \ \ \  ``layout": [x\_top, y\_top, x\_bottom, y\_bottom],
      
\ \ \ \ ``order": 1
      
\ \ \}...]

\}

$<$In-Context Examples$>$
% \end{verbatim}
\end{tcolorbox}
\caption{The simplified prompt template used for our LLM-based Layout Parser. Specific rules are defined for each part of the parsing, and these rules are made comprehensible to the model through In-Context Examples.}
\label{fig:llmprompt}
\end{figure}

\subsection{Dataset Construction}
\label{sup:dataset}
We evaluate LayerBind on two primary tasks: Occlusion Control and General T2I Alignment.
Our prompts are primarily sourced from T2I-CompBench~\cite{huang2023t2i}. However, as this benchmark lacks publicly available layout annotations, we re-annotated the data using our proposed Layout-Parser to ensure a unified input setting. The dataset construction consists of two parts:

\begin{itemize}
    \item \textbf{Layout-annotated T2I-CompBench:} We utilize the \textit{color}, \textit{shape}, \textit{texture}, \textit{2D-spatial}, \textit{numeracy}, and \textit{complex} subsets for general alignment evaluation. For the 3D-spatial subset, which focuses on occlusion, we applied strict filtering. We employed the Layout-Parser to generate bounding boxes and layer orders, followed by manual verification to remove invalid cases (e.g., where regions are fully occluded or exhibit ambiguous 3D relationships). This resulted in 800 high-quality prompts for occlusion evaluation. The sample counts for other subsets remain consistent with the original benchmark.
    
    \item \textbf{BindBench (Complex Occlusion):} The original T2I-CompBench-3D is limited to spatial relationships between only two objects, which is insufficient for evaluating complex occlusion.
    Notably, LaRender~\cite{zhan2025larender} proposed the RealOcc dataset to address similar issues; however, it contains only 60 prompts and is currently not publicly accessible.
    To enable a more comprehensive evaluation, we construct \textbf{BindBench}. We reuse instance categories from T2I-CompBench but recombined them to form complex scenes featuring 3 to 5 overlapping objects. After applying our Layout-Parser for annotation and conducting rigorous manual filtering, we obtain 200 challenging prompts specifically designed to benchmark multi-instance occlusion control.
\end{itemize}

To comprehensively assess LayerBind's performance in terms of layout precision, occlusion accuracy, semantic consistency, and generation quality, we employ the following metrics:

\subsection{Evaluation Metrics}
\label{sup: metrics}
\noindent \textbf{1. UniDet-Depth (Relative Depth Accuracy).}
We utilize the official evaluation script from T2I-CompBench~\cite{huang2023t2i}. This metric employs the UniDet depth estimation model to predict the depth map of the generated image. It then computes the average depth value within the ground-truth bounding boxes of instance pairs to determine if the generated depth order matches the input condition.

\noindent \textbf{2. CLIP Score (Semantic Alignment).}
We assess text-image alignment at both global and local levels using the CLIP ViT-L/14 model:
\begin{itemize}
    \item \textbf{CLIP-G (Global Consistency):} We calculate the cosine similarity between the whole image embedding and the full scene text embedding. This measures how well the overall image captures the global prompt, reflecting both object existence and reasonable spatial composition.
    \item \textbf{CLIP-L (Regional Fidelity):} To evaluate whether specific instances are generated with sufficient regional detail, this metric calculates the similarity between the embedding of the cropped instance region (defined by the layout box) and its corresponding regional prompt.
\end{itemize}

\noindent \textbf{3. L$_{Acc/VQA}$~(Fine-grained Layout Faithfulness).} Following ~\cite{zhang2025creatilayout,zhang2026creatidesign}, which implement VLM-based metrics to evaluate layout faithfulness, we introduce $L_{Acc}$ to measure layout generation accuracy and $L_{attr}$ for fine-grained evaluation of regional attribute fidelity. 
Both metrics are based on VQAScore~\cite{lin2024evaluating} and implemented with Qwen2.5-VL~\cite{bai2025qwen2}.
Specifically, $L_{Acc}$ is computed by questioning each region with the template: ``This image contains \{object class\}?", and $L_{attr}$ is computed by questioning with the detailed regional prompt.
These two metrics are evaluated on the more complex BindBench dataset and offer more discriminative comparisons than the CLIP-L.

\begin{figure}[t]
    \centering
\includegraphics[width=1\linewidth]{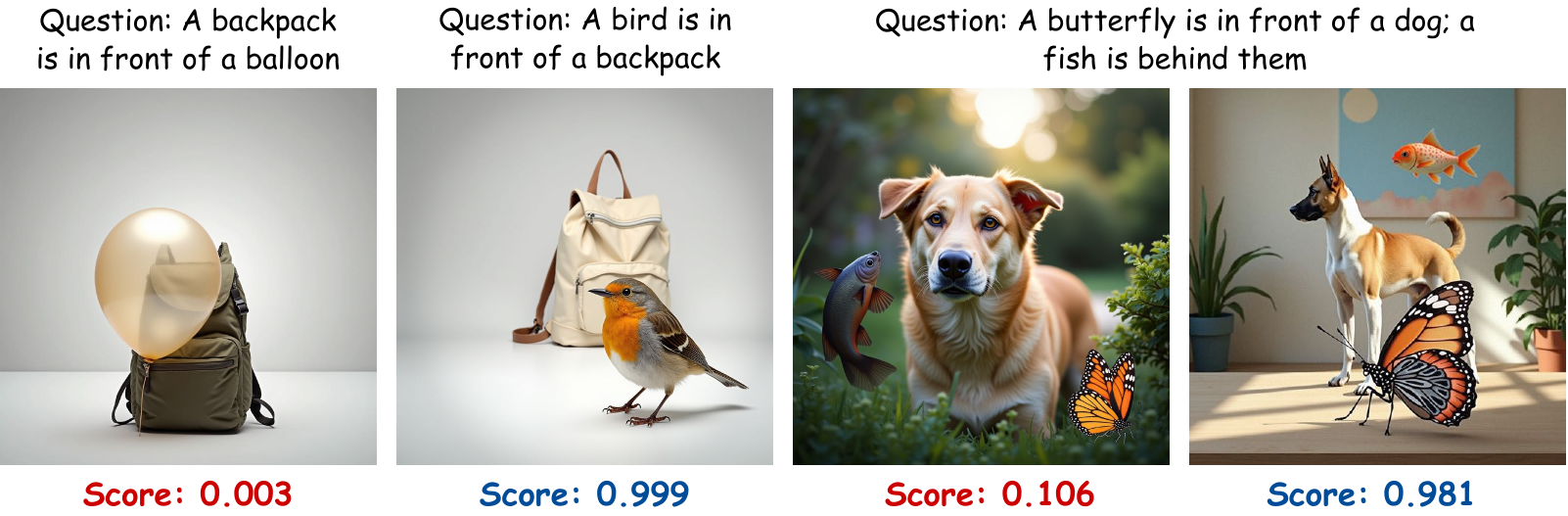}
    \caption{
    Illustration of leveraging QWen-VL2.5~\cite{bai2025qwen2} as a VQA judge for evaluating 3D spatial relationships. It can accurately perceive spatial relationships and score the image content on whether it satisfies the question.
    }
    \label{fig:vqa}
    % \vspace{-5mm}
\end{figure}

\noindent \textbf{4. O$_{VQA}$ (Perceptual Occlusion Success).}
While depth estimation suffices for simple object pairs, it struggles with the complex, multi-instance occlusions found in our BindBench. To address this, we employ the VQAScore~\cite{lin2024evaluating} as a perceptual metric to develop the O$_{VQA}$ metric. Similar to L$_{Acc/VQA}$, we utilize Qwen2.5-VL~\cite{bai2025qwen2} as a visual judge. We feed the generated image and a query regarding the occlusion relationship into the MLLM (as illustrated in Fig.~\ref{fig:vqa}). The model's accuracy serves as a proxy for the human-perceived success rate of occlusion control.

% \noindent \textbf{Metrics Details}.
% \subsection{Metrics Details}

% O$_{VQA}$ and L$_{Acc/VQA}$

\noindent \textbf{5. HPS v2 (Generation Quality).}
Layout control methods often risk degrading image quality (e.g., introducing unnatural lighting or artifacts). To quantify this trade-off, we report the Human Preference Score v2 (HPS)~\cite{wu2023human}. Trained on large-scale human choices, HPS correlates well with human aesthetic judgments. A higher HPS indicates that LayerBind effectively preserves the high-fidelity generation capabilities of the base DiT model.

\begin{table}[t]
\centering
\resizebox{0.8\columnwidth}{!}{%
\begin{tabular}{ccccccc}
\toprule
Model & 1 & 2 & 3 & 4 & 5 & 6 \\
\midrule
FLUX & 18\% & 31\% & 45\% & 60\% & 76\% & 89\% \\
SD3.5 & 24\% & 39\% & 55\% & 73\% & 92\% & 107\% \\
\bottomrule
\end{tabular}
}
\caption{LayerBind's additional inference time when inputting different numbers of regions. Each region occupies 25\% of the image tokens (e.g., 1024 tokens). The inference cost of LayerBind increases linearly with the number of additional tokens.}
\label{table:efficiency}
\vspace{-2mm}
\end{table}

\section{Extended Experiment Analysis}
\label{sup:ablation}

\subsection{Efficiency Analysis}
\label{sup:ablation_efficiency}

In terms of run-time efficiency, the introduction of instance branches inevitably increases the total token count, leading to additional computational overhead.
In Table~\ref{tab:compare1}, we initially reported the inference overhead for standard generation scenarios (based on BindBench cases, involving approximately 40--50\% additional tokens).
To more comprehensively evaluate LayerBind's inference efficiency, Table~\ref{table:efficiency} details the overheads of varying loads, ranging from 1 to 6 input regions (corresponding to a token count increase of 25\% to 150\%).
LayerBind's overhead scales linearly with the number of additional tokens, avoiding the quadratic computational explosion often associated with extended token sequences in Transformers.
This efficiency is attributed to two key design choices: (1) branch tokens are active only during the limited early initialization steps, and (2) our local update calculation employs an block-wise strategy rather than full-sequence calculation.
In summary, LayerBind maintains a highly practical trade-off, achieving precise control with manageable computational costs.

\begin{figure}[h]
    \centering
\includegraphics[width=1\linewidth]{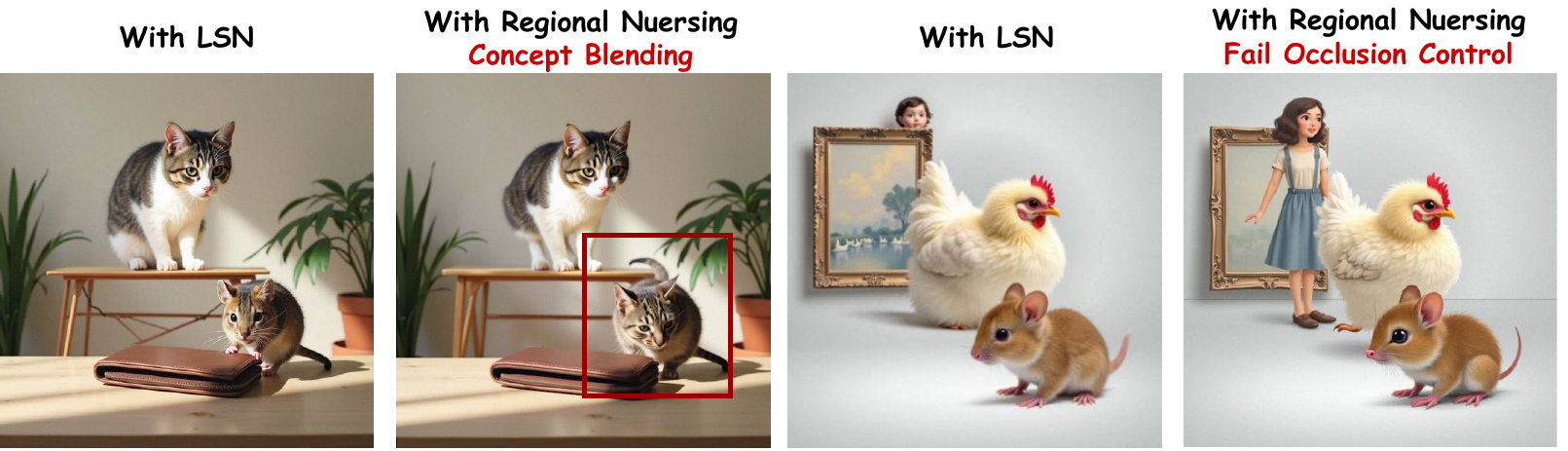}
    \caption{
    The illustration of the effectiveness of the proposed LSN and naive regional prompting~\cite{chen2024training} strategies. Without layer-wise updates, errors such as concept blending and failure in occlusion control may occur.
    }
    \label{fig:ablation3}
    \vspace{-5mm}
\end{figure}

\subsection{Layer-wise Nursing vs. Regional Prompting}

A straightforward alternative to our Layer-wise Semantic Nursing (LSN) is standard regional prompting~\cite{chen2024training}, which injects semantics without explicit layer-wise compositing.
We observe that for layouts with spatially disjoint instances (i.e., no complex occlusion), Regional Prompting effectively refines instance details.
However, as illustrated in Fig.~\ref{fig:ablation3}, it fails in two critical scenarios:

\begin{itemize}
    \item Concept Blending: Without the explicit isolation, attention leakage may occur between regions. This can lead to concept blending, where semantics blend across boundaries even without explicit visual overlap.
    \item Occlusion Failure: During the nursing phase, the initialized latent representations often lack sufficiently distinct semantic boundaries to maintain occlusion naturally. Without layer-wise updates to strictly enforce visibility ordering, standard regional prompting fails to ensure that the foreground robustly overwrites the background. Consequently, the pre-established occlusion relationships degrade or vanish in the final output.
\end{itemize}

Therefore, we conclude that the proposed LSN is a more robust strategy, essential for maintaining both semantic and the correct occlusion order.

\begin{figure}[t]
    \centering
    \includegraphics[width=1\linewidth]{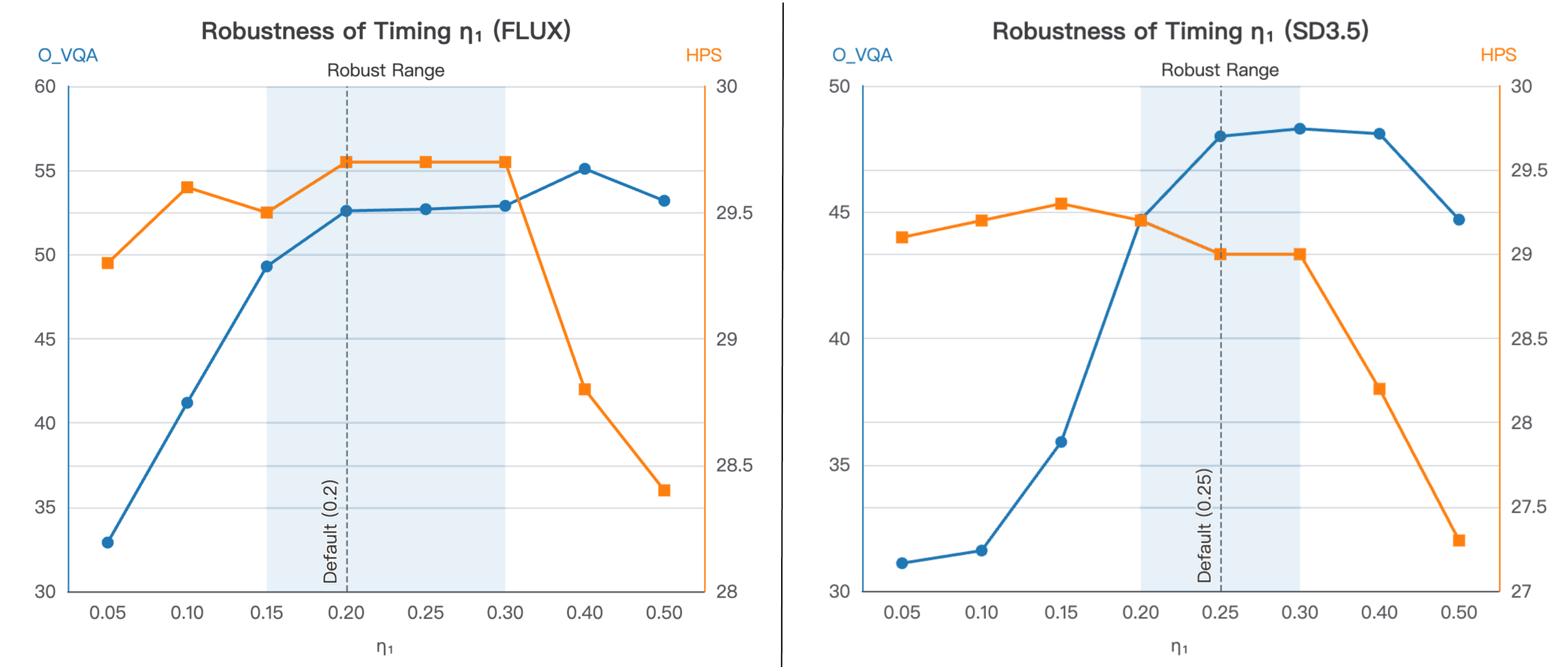}
    \vspace{-5mm}
    \caption{
        The performance is stabilized across a robust $\eta_1$ range, while fine-tuning $\eta_1$ can further optimize specific cases.
    }
    \label{fig:plot}
    % \vspace{-5mm}
\end{figure}

\begin{figure}[t]
    \centering
    \includegraphics[width=1\linewidth]{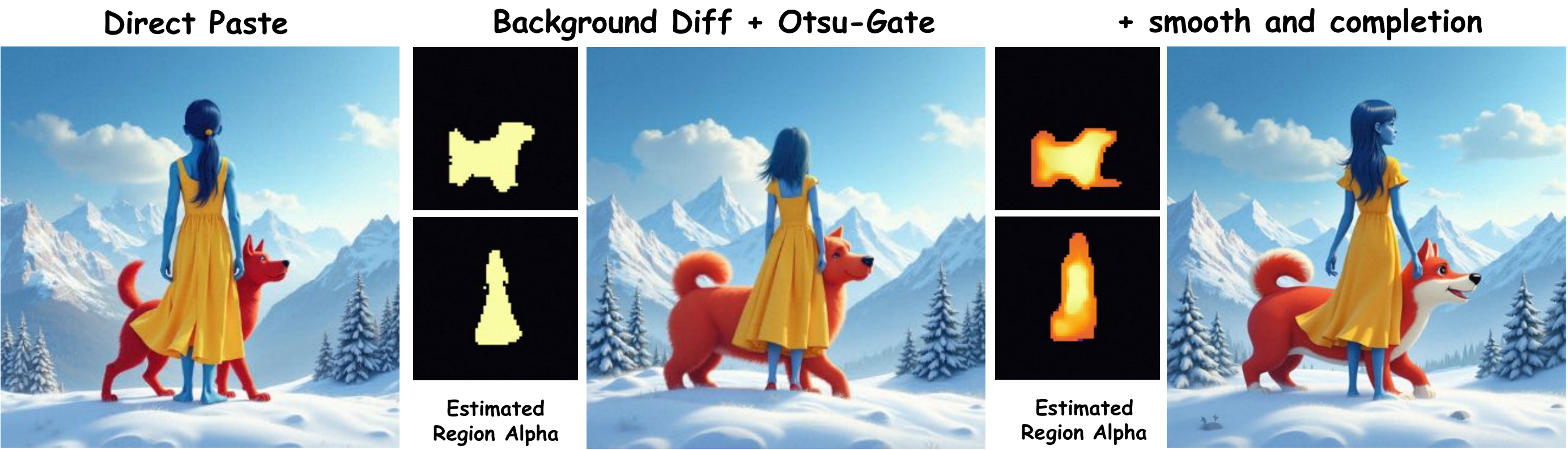}
    % \vspace{-5mm}
    \caption{
        The visualization of the estimated alpha mask and the effectiveness of different branch blending strategies.
    }
    \label{fig:blend_alpha}
    \vspace{-5mm}
\end{figure}

\subsection{Effect of $\eta_1$ and $\eta_2$}

The selection of $\eta_1$ and $\eta_2$ is critical for LayerBind's performance.
First, $\eta_1$ is the decisive factor for generation success. As illustrated in Fig.~\ref{fig:ablation2}, an excessively low $\eta_1$ fails to impose sufficient spatial constraints, while an overly high value causes significant stylistic or layout over-decopled between the instance and the background (Fig.~\ref{fig:ablation3}).
As shown in Fig.~\ref{fig:plot}, $\eta_1 = 0.25$ serves as a robust default for both FLUX and SD3.5, though fine-tuning within $[0.1, 0.3]$ can further optimize specific cases.  Regarding $\eta_2$, its role includes semantic detail refinement with layout maintenance. For simple instances where structural preservation is the priority, $\eta_2 = 0.5$ suffices. However, for instances with complex attribute details, we recommend increasing $\eta_2$ to 0.7 to ensure faithful semantic details.

\subsection{Effectiveness of Branch Blending}

The Branch Blending mechanism works as shown in Fig.~\ref{fig:blend_alpha}, only applied to the occluded instances region, which is an optional step for enhancing the instance edge quality. 
We found that in most cases, as long as the bottom-layer instances have sufficient unoccluded parts, direct paste can maintain the occlusion, while further blending strategies further improve the generation quality of both the instances and the overall scene.

% Branch Blending机制的作用如Fig.~\ref{fig:blend_alpha}所示，我们将其视为一种用于提升instance边缘质量的optional module. 我们发现在大部分case中，只要靠近镜头的instance不完全遮挡背景（背景instance仍有充分的暴露部分），直接粘贴在大多时候也能够成功维持遮挡。加入基于背景差分以及边缘smooth的alpha estimation后，能够进一步提升场景与instance的生成质量。

% As introduced in Sec.~\ref{sec:layer_blend} and ~\ref{sup:method}, we propose an optional branch blending mechanism.
\begin{figure*}[t]
    \centering
\includegraphics[width=1\linewidth]{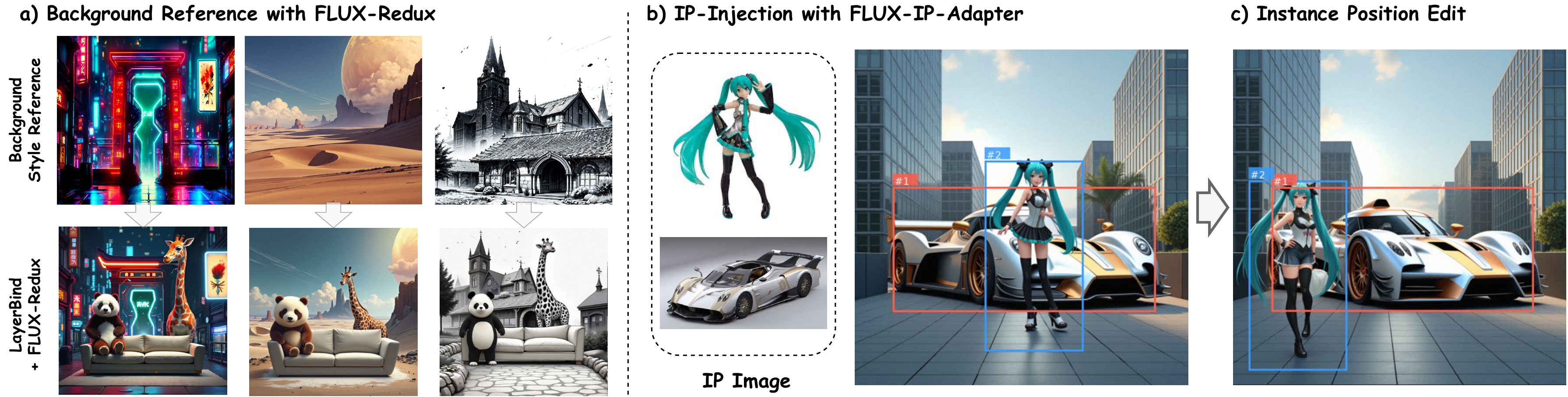}
    \caption{The proposed LayrBind can be integrated with external visual adapters. a) With the FLUX-Redux plugin, LayerBind can directly generate the background by referencing an existing image. b) With IP-adapter~\cite{ye2023ip}, LayerBind can inject visual references into regions. c) LayerBind supports instance position editing.
    }
    \label{fig:more_app}
    % \vspace{-5mm}
\end{figure*}

\section{Extended Applications}
\label{sup:app}

\subsection{Implementation of Composited Image Editing}
As illustrated in Fig.~\ref{fig:applications}, LayerBind supports multi-region, multi-instruction composited image editing.
This capability stems from LayerBind's inherent ability to spawn and merge instance branches onto any arbitrary background generation trajectory.
Specifically, the editing pipeline proceeds as follows:

\begin{itemize}
    \item First, we obtain the denoising trajectory of the image. For synthesized images, we can directly use the original prompt and initial noise; for real images, inversion methods~\cite{wang2024taming} can be used to retrieve the denoising trajectory.

    \item Then, a key distinction from our standard layout control (which branches at pure noise $t=T$) is the timing of branch creation. For editing tasks, to ensure optimal structural consistency with the original image, we instantiate the edit branches after the first denoising step rather than at initialization.

    \item Finally, each edit region undergoes an independent branch guided by its specific instruction via Eq.~\ref{eq:contextual_update}, functioning analogously to parallel multi-region image inpainting. For fusion, we employ a larger initialization ratio ($\eta_1 \approx 0.4\text{-}0.5$). At this stage, the edited content and structure are firmly established; completing the subsequent standard denoising process yields the final edited result.
\end{itemize}

We plan to provide more extensive analysis and examples of this application in future iterations of this work.

\subsection{Compatibility with External Adapters}

Since LayerBind relies solely on attention mechanisms, it is inherently compatible with most external adapters.
We exemplify this capability in Fig.~\ref{fig:more_app} using two popular tools: the FLUX Redux adapter and IP-Adapter. First, LayerBind can utilize the Redux adapter as a substitute for the textual background prompt. This allows the global scene style to be initialized directly from a reference image while maintaining LayerBind's structural control. Then, LayerBind integrates seamlessly with IP-Adapter. By injecting visual priors from the IP-Adapter into specific instance regions, LayerBind simultaneously governs both the spatial layout and the specific visual reference.

\subsection{Generation with Transparent Instances}

\begin{figure}[t]
    \centering
\includegraphics[width=1\linewidth]{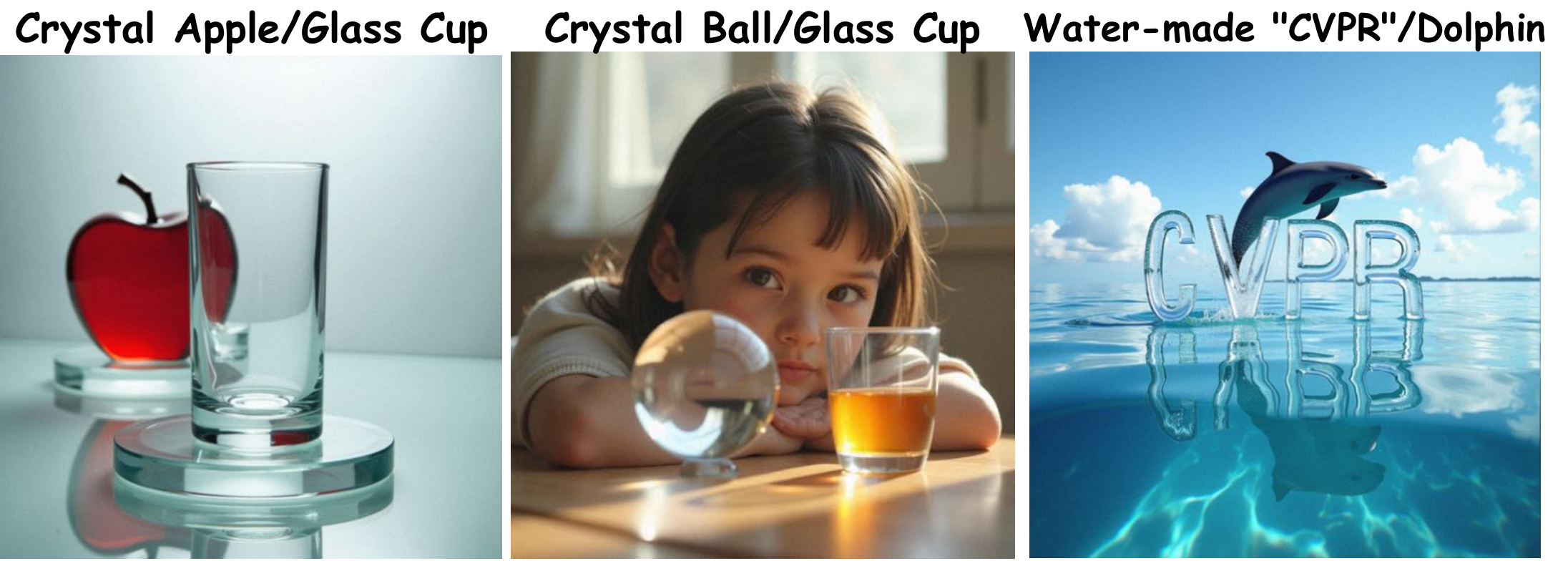}
    \caption{
    Visualization of generating transparent instances.
    }
    \label{sup:transparent}
    % \vspace{-5mm}
\end{figure}

Fig.~\ref{sup:transparent} showcases some generation results with transparent instances/reflection, LayerBind can naturally handle transparent object generation without occlusion scenarios. 
When there is occlusion, since transparent objects are difficult to estimate region alpha using background differences, we directly use direct paste to handle the occluded regions.
We found that under the subsequent nursing mechanism, LayerBind can also handle occlusion relationships between transparent instances.

\subsection{Implementation of Position Editing}

As shown in Fig.~\ref{fig:more_app}~(c), by making the implementation adjustments, LayerBind can handle instance position editing. 
The Branch mechanism of LayerBind is built on a shared noisy latent of the regions; therefore, modifying the region coordinates will change the generation trajectory.
If it wishes to preserve the originally generated instance features during position editing, alpha estimation should be performed on the instance region before blending it into the new region, rather than directly changing the region coordinates of the instance.

\subsection{Complex Scene Generation~(>10 Instances)}

\begin{figure}[t]
    \centering
\includegraphics[width=1\linewidth]{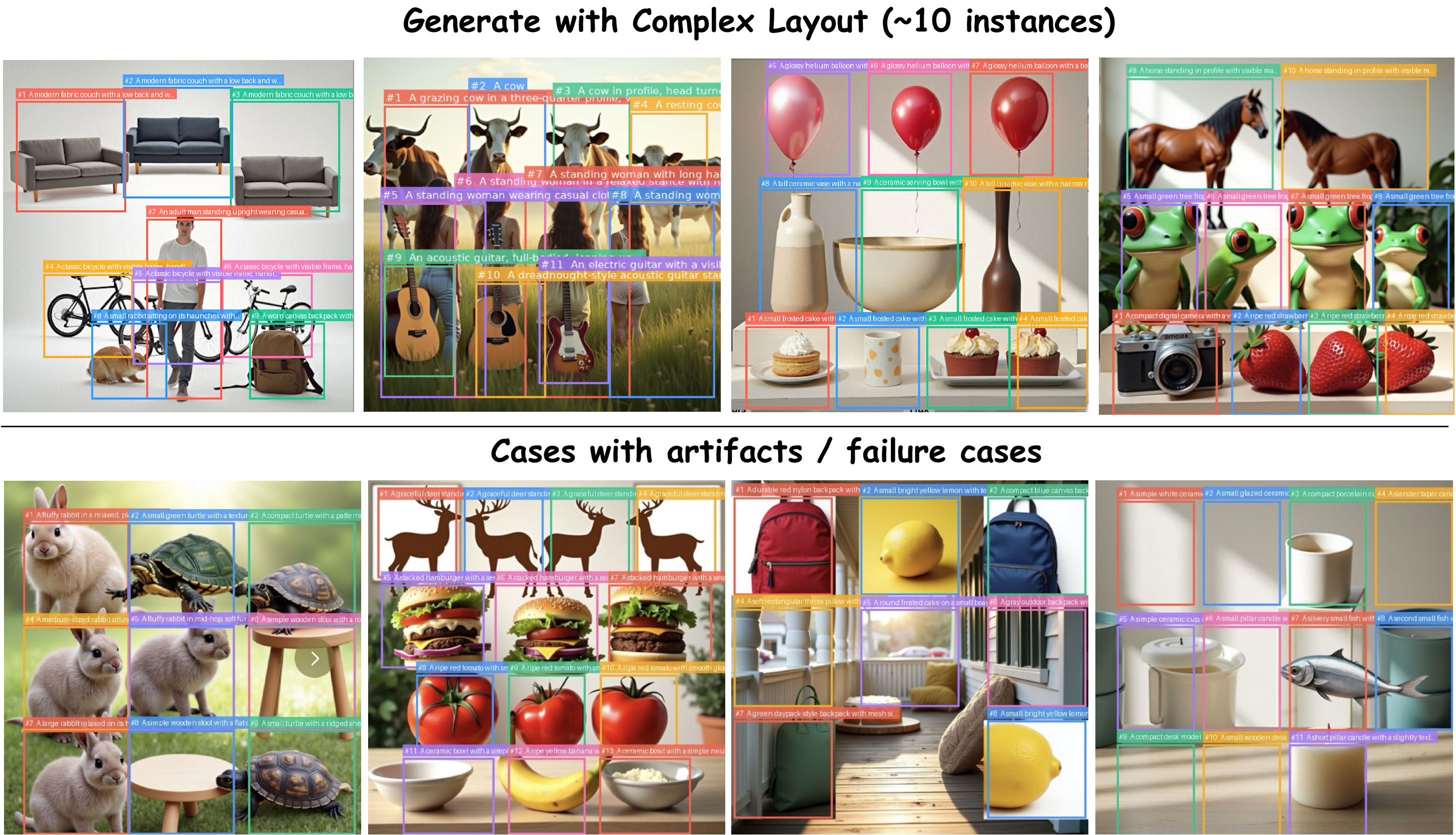}
    \caption{
    % Visualization of generating with complex layouts.
    LayerBind successfully handles rational complex spatial relationships, while failing in counterfactually incorrect layout arrangements.
    }
    \label{sup:complex}
    % \vspace{-5mm}
\end{figure}

In Fig.~\ref{sup:complex}, we present the results of using LayerBind to generate more instances of complex layouts.
We find that when the input layout itself is spatially logical (i.e., all objects are correctly positioned in the background in a logically reasonable manner), LayerBind is able to accurately handle complex arrangements with more than 10 instances. 
However, with such a large number of instances, most of the input cases are counterfactual. Since LayerBind is a training-free method that only uses the context sharing ability inherent in the model itself, it struggles to handle generation scenarios outside of its training data distribution, which can lead to artifacts or binding failures.
\vspace{10mm}

\section{Limitations}
\label{sup:limitation}

\begin{figure}[t]
    \centering
\includegraphics[width=1\linewidth]{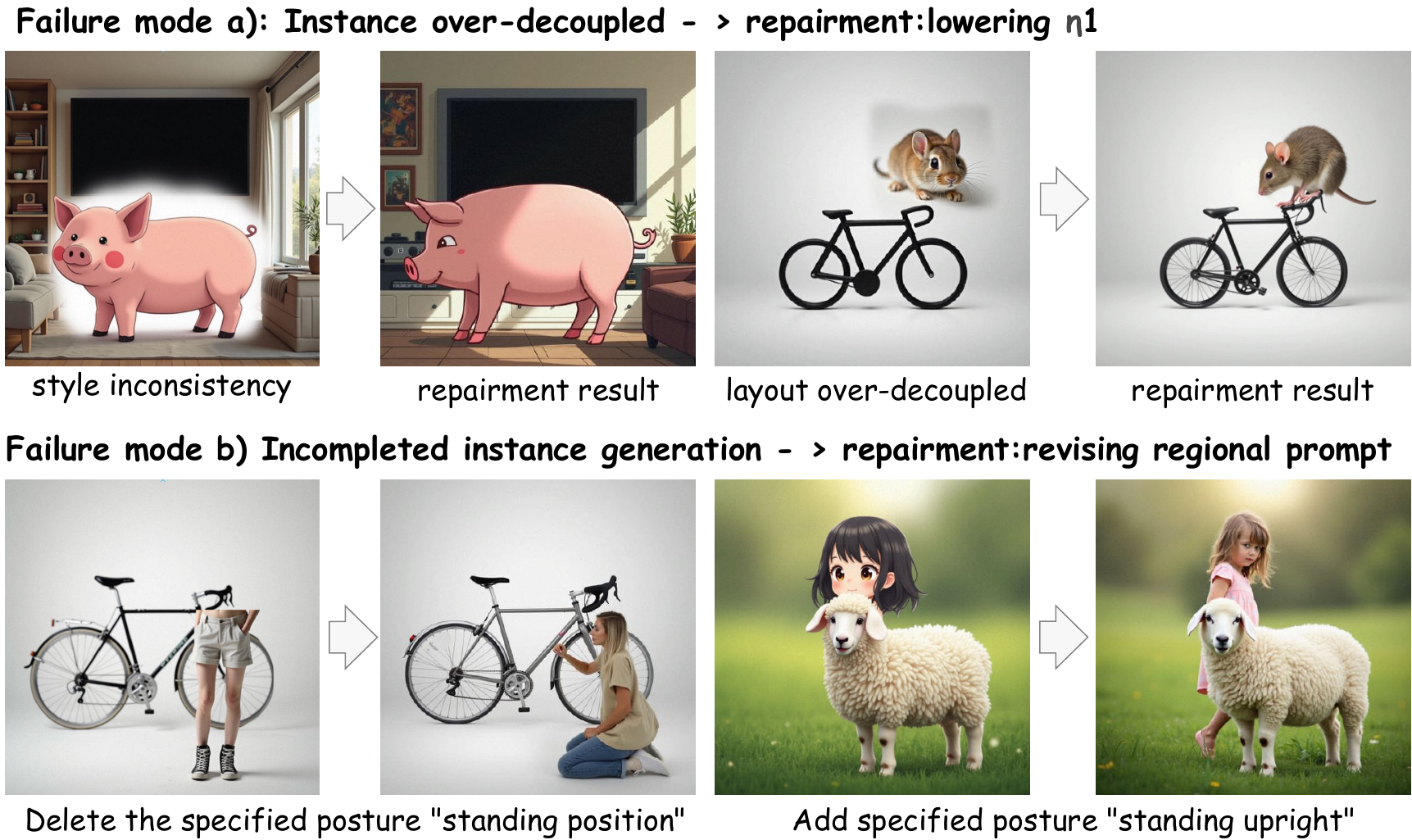}
    \caption{
    Illustration of typical failure cases and repair measures.
    }
    \label{fig:failure}
    % \vspace{-5mm}
\end{figure}

Fig.~\ref{fig:failure} illustrates representative failure cases encountered by LayerBind, which can be summarized as follows:
\begin{itemize}
    \item \textbf{Instance-Background over-decoupling:} In some results, we observe a stylistic or structural detachment between the instance and the background. This typically arises when $\eta_1$ is set too high, leaving the subsequent nursing phase insufficient capacity to restore global harmony. Reducing $\eta_1$ (sacrificing a degree of rigid layout control) effectively mitigates this issue.
    
    \item \textbf{Incomplete instance generation:} The successful instance generation is dependent on the alignment between its regional prompt and its spatial location. For example, as shown in the figure, depending on the position of the human, adjusting their poses in prompts can avoid incomplete generation.
\end{itemize}

Beyond the visualized cases, we note that LayerBind struggles with ``Dense Layout" scenarios common in traditional Layout-to-Image benchmarks~\cite{zheng2023layoutdiffusion,zhang2025creatilayout,xiang2025instanceassemble,wu2025hybrid}. This limitation stems from our core design: decoupling background and instance generation makes it challenging to maintain holistic consistency in highly cluttered scenes. Nevertheless, we maintain that LayerBind's primary strength lies in \textit{customized generation} scenarios as a training-free controller, rather than dense scene synthesis.
Future work could explore integrating LayerBind's mechanics with model fine-tuning strategies to achieve stronger global coherence while retaining precise regional and occlusion control.

\section{More Visualizations}
\label{sup:vis}

\begin{figure*}[t]
    \centering
    \includegraphics[width=1\textwidth]{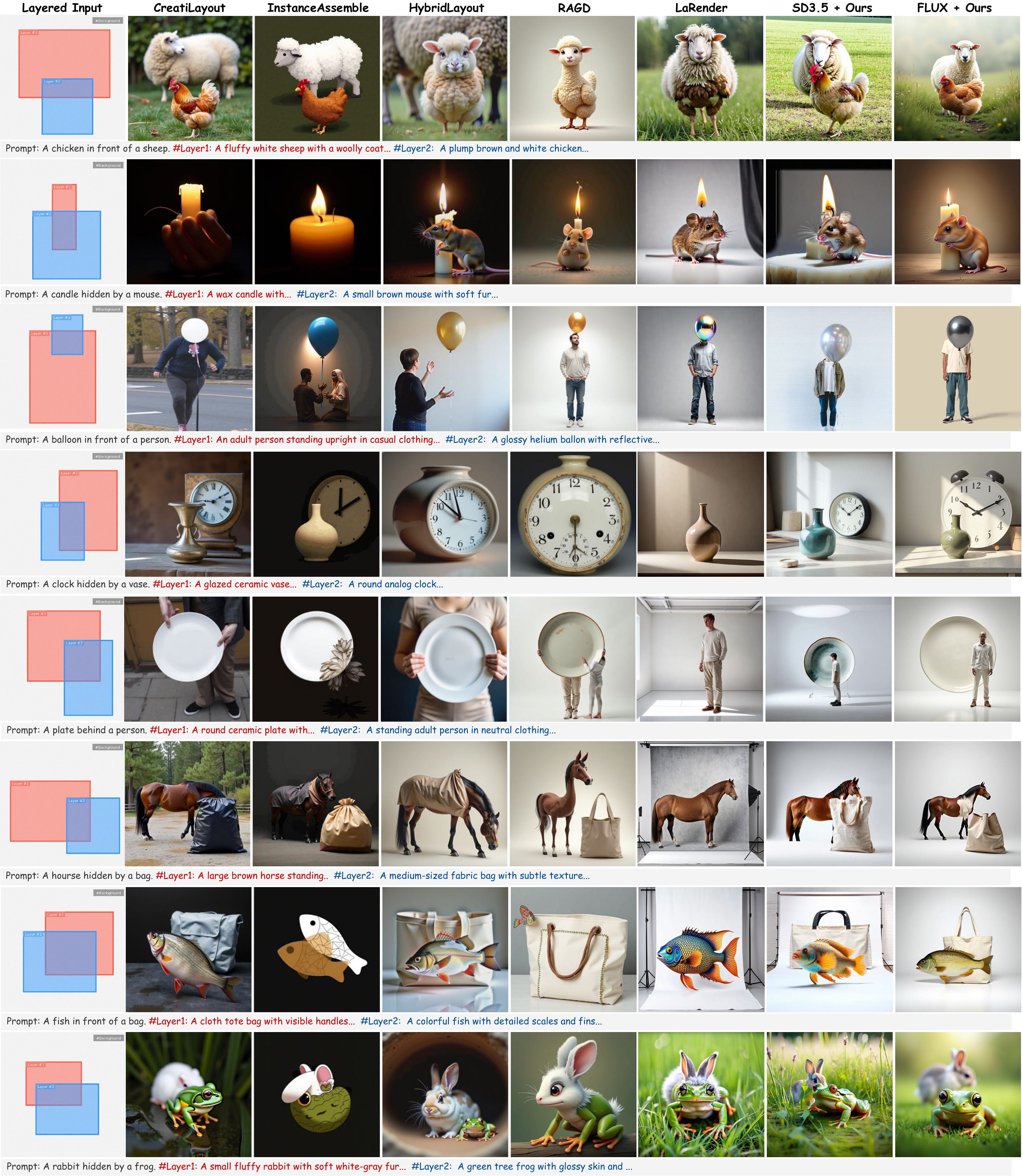}
    \captionof{figure}{Visualization of occlusion control abilities on T2ICompBench-3D dataset.} 
    \label{fig:compare_t2i3d}
    \vspace{-5mm}
    % \vspace{10mm}
\end{figure*}

\begin{figure*}[t]
    \centering
    \includegraphics[width=1\textwidth]{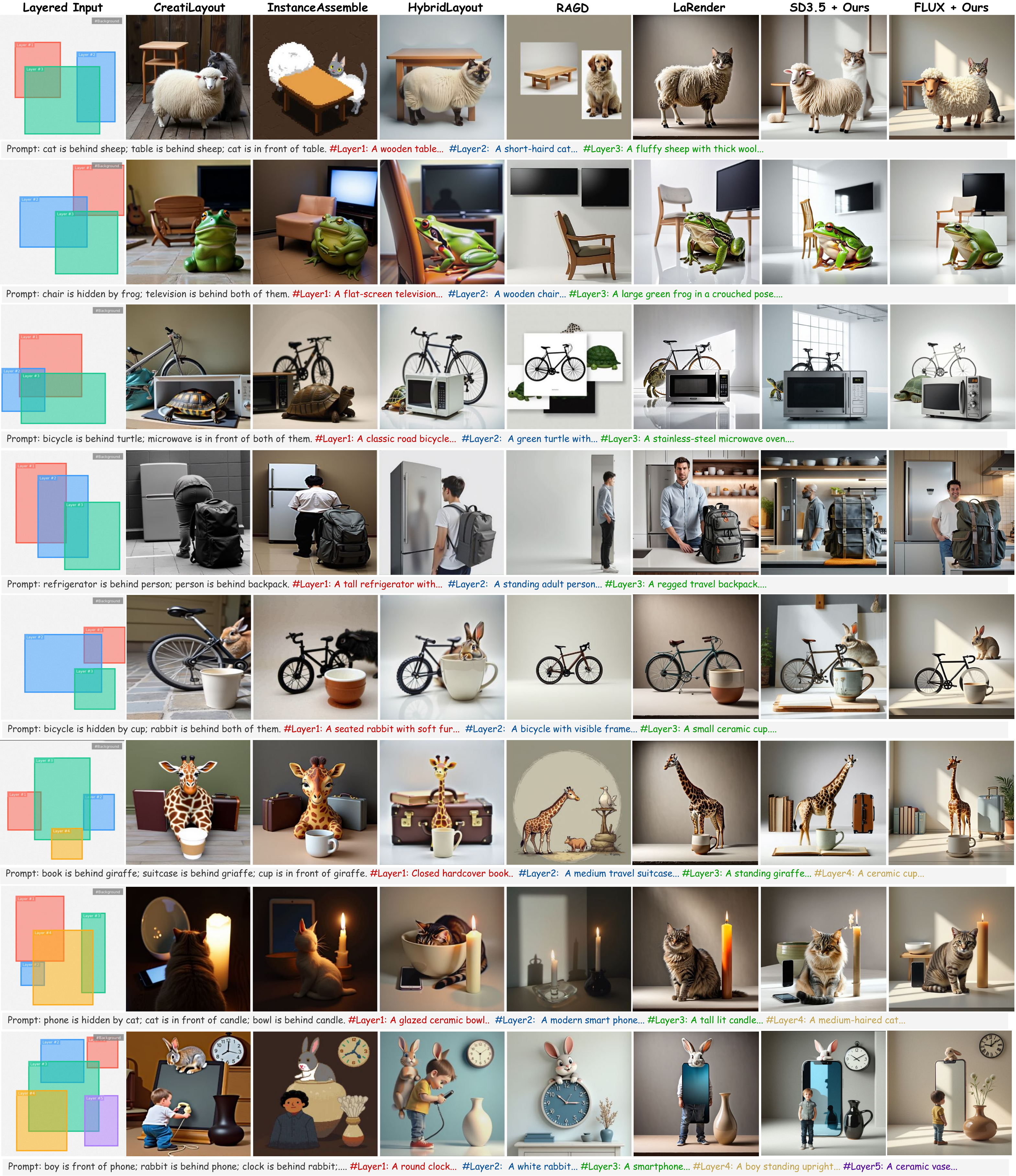}
    \captionof{figure}{Visualization of occlusion control abilities on BindBench dataset.} 
    \label{fig:compare_bindbench}
    \vspace{-5mm}
    % \vspace{10mm}
\end{figure*}

\begin{figure*}[t]
    \centering
    \includegraphics[width=1\textwidth]{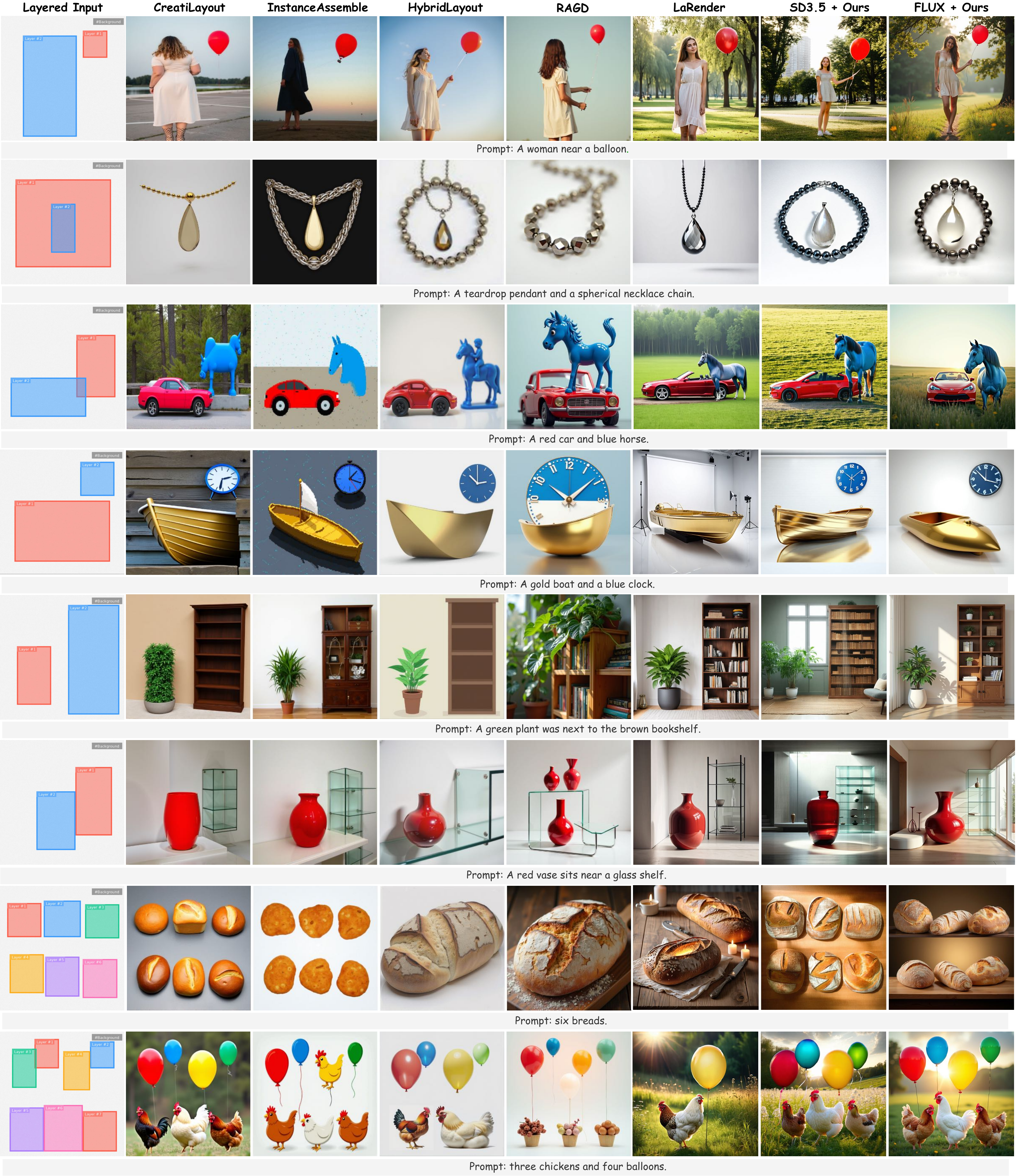}
    \captionof{figure}{Visualization of T2I alignment tasks on T2ICompBench dataset.} 
    \label{fig:compare_t2i}
    \vspace{-5mm}
    % \vspace{10mm}
\end{figure*}

Finally, we show cases of more visualization results on the evaluated datasets, in Figs.~\ref{fig:compare_t2i3d},~\ref{fig:compare_bindbench}, and~\ref{fig:compare_t2i}.

% \section{Rationale}
% \label{sec:rationale}
% % 
% Having the supplementary compiled together with the main paper means that:
% % 
% \begin{itemize}
% \item The supplementary can back-reference sections of the main paper, for example, we can refer to \cref{sec:intro};
% \item The main paper can forward reference sub-sections within the supplementary explicitly (e.g. referring to a particular experiment); 
% \item When submitted to arXiv, the supplementary will already included at the end of the paper.
% \end{itemize}
% % 
% To split the supplementary pages from the main paper, you can use \href{https://support.apple.com/en-ca/guide/preview/prvw11793/mac#:~:text=Delete%20a%20page%20from%20a,or%20choose%20Edit%20%3E%20Delete).}{Preview (on macOS)}, \href{https://www.adobe.com/acrobat/how-to/delete-pages-from-pdf.html#:~:text=Choose%20%E2%80%9CTools%E2%80%9D%20%3E%20%E2%80%9COrganize,or%20pages%20from%20the%20file.}{Adobe Acrobat} (on all OSs), as well as \href{https://superuser.com/questions/517986/is-it-possible-to-delete-some-pages-of-a-pdf-document}{command line tools}.


\end{document}